\documentclass[nohyperref]{article}

\usepackage{enumitem} 
\usepackage{color, colortbl}
\usepackage{xcolor}
\usepackage{pifont}

\usepackage{graphicx}
\usepackage{epstopdf}
\usepackage{subcaption}
\usepackage{pgfplots}
\usepackage{pgfplotstable}
\usepackage{hyperref}
\usepackage{comment}
\usepackage{amsmath}
\usepackage{bm}
\usepackage{subcaption}
\usepackage{xcolor}

\usepackage[capitalise]{cleveref}
\usepackage{amssymb}
\usepackage{siunitx}
\usepackage{listings}
\usepackage[utf8]{inputenc}

\usepackage{booktabs}  
\usepackage{tabu}
\usepackage{array}
\usepackage{multirow}
\usepackage{diagbox}
\usepackage{amsthm}
\usepackage{wrapfig}
\usepackage{placeins}

\usepackage{algorithm,lipsum,changepage}
\usepackage{shortcuts_ln}

\usepackage{balance}

\usepackage{enumitem}
\usepackage{microtype}
\usepackage{calc}
\usepackage{ifthen}
\usepackage{hyphenat}
\usepackage{fancyhdr}
\usepackage{color}
\usepackage{algorithmic}
\usepackage{natbib}
\usepackage{eso-pic} 
\usepackage{forloop}

\usepackage{hyperref}       
\usepackage{url}     
\usepackage{booktabs}       
\usepackage{amsfonts}       
\usepackage{nicefrac}       
\usepackage{microtype}      
\usepackage{xcolor}         
\usepackage{mathabx}
\usepackage{microtype}
\usepackage{graphicx}

\usepackage{booktabs} 

\usepackage{hyperref}

\usepackage[accepted]{icml2024}

\usepackage{amsmath}
\usepackage{amssymb}
\usepackage{mathtools}
\usepackage{amsthm}

\theoremstyle{plain}
\newtheorem{theorem}{Theorem}[section]
\newtheorem{proposition}[theorem]{Proposition}

\theoremstyle{definition}

\theoremstyle{remark}

\usepackage[textsize=tiny]{todonotes}

\icmltitlerunning{Vanilla Bayesian Optimization Performs Great in High Dimensions\textbf{}}

\begin{document}

\twocolumn[
\icmltitle{Vanilla Bayesian Optimization Performs Great in High Dimensions}



\icmlsetsymbol{equal}{*}

\begin{icmlauthorlist}
\icmlauthor{Carl Hvarfner}{yyy}
\icmlauthor{Erik O. Hellsten}{yyy,comp}
\icmlauthor{Luigi Nardi}{yyy,comp}
\end{icmlauthorlist}

\icmlaffiliation{yyy}{Lund University, Lund, Sweden}
\icmlaffiliation{comp}{DBtune, Malmö, Sweden}

\icmlcorrespondingauthor{Carl Hvarfner}{carl.hvarfner@cs.lth.se}

\icmlkeywords{Machine Learning, ICML}

\vskip 0.3in
]




\newcommand{\data}[0]{\mathcal{D}}
\newcommand{\hps}[0]{\bm{\theta}}
\newcommand{\yx}[0]{y_{\bm{x}}}
\newcommand{\yxf}[0]{y_{\bm{x}|*}}
\newcommand{\opt}[0]{\bm{\bigast}}

\newcommand{\jes}[0]{\texttt{JES}}
\newcommand{\KG}[0]{\texttt{KG}}
\newcommand{\acqpi}[0]{\texttt{PI}}
\newcommand{\sal}[0]{\texttt{SAL}}
\newcommand{\ei}[0]{\texttt{EI}}
\newcommand{\pes}[0]{\texttt{PES}}
\newcommand{\es}[0]{\texttt{ES}}
\newcommand{\mes}[0]{\texttt{MES}}
\newcommand{\fitbo}[0]{\texttt{FITBO}}
\newcommand{\scbo}[0]{\texttt{SCoreBO}}
\newcommand{\imspe}[0]{\texttt{IMSPE}}
\newcommand{\alm}[0]{\texttt{ALM}}
\newcommand{\balm}[0]{\texttt{BALM}}
\newcommand{\bald}[0]{\texttt{BALD}}
\newcommand{\bqbc}[0]{\texttt{BQBC}}
\newcommand{\qbmgp}[0]{\texttt{QBMGP}}
\newcommand{\logei}[0]{\texttt{LogEI}}
\newcommand{\xopt}[0]{\bm{x}^*}
\newcommand{\cand}[0]{\bm{x}_*}
\newcommand{\obs}[0]{\bm{X}}
\newcommand{\fopt}[0]{f^*}
\newcommand{\setn}[0]{{\bm{X}_n}}
\printAffiliationsAndNotice{}

\begin{abstract}
High-dimensional problems have long been considered the Achilles' heel of Bayesian optimization. Spurred by the curse of dimensionality, a large collection of algorithms aim to make it more performant in this setting, commonly by imposing various simplifying assumptions on the objective. In this paper, we identify the degeneracies that make vanilla Bayesian optimization poorly suited to high-dimensional tasks, and further show how existing algorithms address these degeneracies through the lens of lowering the model complexity. Moreover, we propose an enhancement to the prior assumptions that are typical to vanilla Bayesian optimization, which reduces the complexity to manageable levels without imposing structural restrictions on the objective. Our modification - a simple scaling of the Gaussian process lengthscale prior with the dimensionality - reveals that standard Bayesian optimization works drastically better than previously thought in high dimensions, clearly outperforming existing state-of-the-art algorithms on multiple commonly considered real-world high-dimensional tasks.
\end{abstract}

\section{Introduction}

In Bayesian optimization, \emph{complexity} and \emph{dimensionality} are intrinsically interlinked — the higher the problem dimensionality, the harder it is to optimize. 
The exuberance of space, and large distance between observations, makes the size of high-variance regions along the boundary of the search space exponentially large~\cite{maju2021hdsurvey, binois2022hdsurvey}.
Moreover, the growing number of parameters of the Gaussian Process (GP) surrogate in relation to the number of observations makes accurate modeling of the problem at hand exceedingly difficult. In recent years, the effort to create methods that achieve efficient Bayesian optimization (BO) in high dimensions has been substantial, making it one of the most frequently addressed challenges in the BO research community~\cite{kandasamy-icml15a, wang2016bayesian, nayebi2019framework, eriksson2019turbo, pmlr-v161-eriksson21a, papenmeier2022increasing, papenmeier2023bounce, pmlr-v202-ziomek23a}.

While approaches are plentiful and diverse, they all share a common characteristic: they employ restrictions on the objective which reduces its a-priori \textit{assumed complexity} by contracting the search space. This in turn decreases distances between data points and prospective queries, increasing their correlation, thus making GP inference more informative. Assuming a degree of complexity which enables meaningful correlation is essential to efficiently optimize problems of \textit{any} dimensionality. Nevertheless, the high-complexity, low-correlation issue presents itself most clearly in the high-dimensional setting. 

In this paper, we hypothesize that the shortcomings of vanilla BO in high dimensions are strictly a consequence of the complexity assumptions imposed on the objective. To that end, we view existing high-dimensional BO (HDBO) approaches through the lens of model complexity, which arises from their structural assumptions. Thereafter, we modify standard BO to follow a similarly complexity reduced structure, simply by appropriately scaling the lengthscale prior of the GP kernel. Consequently, we effectively circumvent the well-established Curse of Dimensionality (CoD) without introducing any of the conventional structural restrictions on the objective that are prevalent in HDBO. We demonstrate that standard BO works drastically better than previously thought for high-dimensional tasks, outclassing existing high-dimensional BO algorithms on a wide range of real-world problems. Further, we aim to shed light on the inner workings of the BO machinery and why minimal changes in assumptions yield a dramatic increase in performance. The result is a performant \textit{vanilla} BO algorithm for dimensionalities well into the thousands.

Formally, we make the following contributions:
\begin{enumerate}
    \item We demonstrate the crucial difference between dimensionality and complexity in BO, highlighting the failure modes related to high \textit{assumed complexity} and relate existing HDBO classes to a reduction in complexity. 
    \item We prove that when the model is uninformed, ~\ei{} will \textit{not} exihibit exploratory behavior along the boundary, contrasting claims of~\cite{swersky2017improving, oh18bock}.
    \item We propose a plug-and-play enhancement to the vanilla BO algorithm that reduces the assumed complexity to enable high-dimensional optimization, and extensively validate it across a wide spectrum of dimensionalities. Results show that vanilla BO works significantly better for high-dimensional problems than previously imagined, substantially outperforming state-of-the-art HDBO methods on a wide range of real-world tasks.
\end{enumerate}

\section{Background}

In this section, we review the background related to Gaussian processes and Bayesian optimization. We outline the maximal information gain (MIG) as a measure of problem complexity, and the model-level choices that impact the a-priori assumed problem complexity, to subsequently explore pitfalls of vanilla BO for high-complexity tasks in Sec~\ref{sec:issues}.

\subsection{Gaussian Processes}
The Gaussian process (GP) has become the model class of choice in most BO applications. The GP provides a distribution over functions $\hat{f}\sim \mathcal{GP}(m(\cdot), k(\cdot, \cdot))$ fully defined by the mean function $m(\cdot)$ and the covariance function $k(\cdot, \cdot)$. Under this distribution, the value of the function $\hat{f}(\bm{x})$, at a given location $\bm{x}$, is normally distributed with a closed-form solution for the mean $\mu(\bm{x})$ and variance $\sigma^2(\bm{x})$. We model a constant mean, so that the dynamics are fully determined by the covariance function $k(\cdot, \cdot)$. 

To account for differences in variable importance, each dimension is individually scaled using lengthscale hyperparameters $\ell_i$. This is commonly referred to as Automatic Relevance Determination (ARD)~\cite{NIPS1995_7cce53cf}. For $D$-dimensional inputs $\bm{x}$ and $\bm{x}'$, the distance $r(\bm{x}, \bm{x}')$ is subsequently computed as $r^2 = \sum_{{i}=1}^D (x_{i} - x_{i}')^2 / \ell_{i}^2$. Along with the signal variance $\sigma_f$ and \emph{noise variance} $\sigma_\varepsilon^2$, $\hps =\{\bm{\ell}, \sigma^2_\varepsilon, \sigma^2_f \}$ comprise the set of hyperparameters that are conventionally learned, with a possible addition of a learnable constant mean $c$~\cite{balandat2020botorch, snoek-nips12a}. The likelihood surface $p(\hps|\data)$ for the GP hyperparameters is typically highly multi-modal \cite{rasmussen-book06a, yao2020stacking} and desirable hyperparameters are conventionally found by MAP estimation, where a hyperprior is set on the kernel hyperparameters $\hps$. While often overlooked, the choice of hyperprior can greatly impact the performance of a BO algorithm in practice, particularly in non-conventional problem settings~\cite{pmlr-v161-eriksson21a, pmlr-v80-baptista18a, pmlr-v139-rothfuss21a, rothfuss2021fpacoh, hvarfner2023selfcorrecting}.

\subsection{Bayesian Optimization}\label{sec:bo}
We aim to find a maximizer $\bm{x}^* \in \argmax_{\bm{x}\in \mathcal{X}} f(\bm{x})$ of the black-box function $f(\bm{x}):\mathcal{X} \rightarrow\mathbb{R}$, over the $D$-dimensional input space $\mathcal{X}=[0, 1]^D$.
We assume that $f$ can only be observed point-wise and that the observations are perturbed by Gaussian noise, $y(\bm{x}) = f(\bm{x})+\varepsilon_i$ with $\varepsilon_i\sim \mathcal{N}(0, \sigma_\varepsilon^2)$.

The \textit{acquisition function} uses the surrogate model to quantify the utility of a point in the search space. Acquisition functions employ a trade-off between exploration and exploitation, typically using a greedy heuristic to do so. Most common is the Expected Improvement (\ei{})~\cite{jones-jgo98a, bull-jmlr11a} and its numerically stable, easy-to-optimize adaptation \logei{}~\cite{ament2023unexpected}. Another acquisition function which uses similar heuristics is the Upper Confidence Bound~\cite{Srinivas_2012, srninivas-icml10a}.

\begin{figure*}[htbp]
    \centering
\begin{minipage}[b]{0.744\textwidth}
    \includegraphics[width=\textwidth]{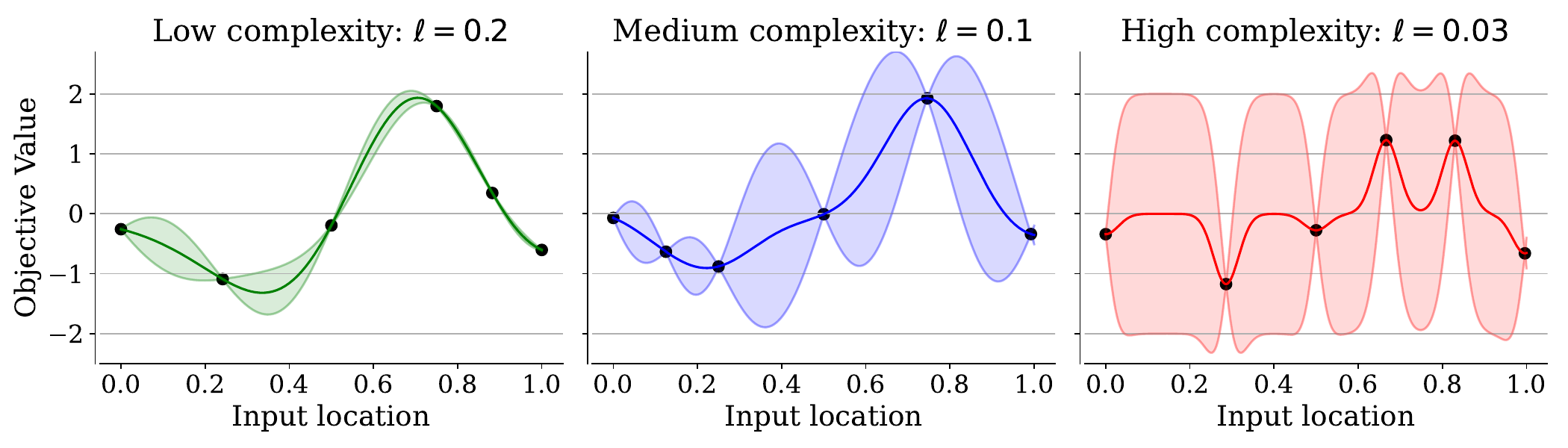}
\end{minipage}
\hfill
\begin{minipage}[b]{0.248\textwidth}
    \includegraphics[width=\linewidth]{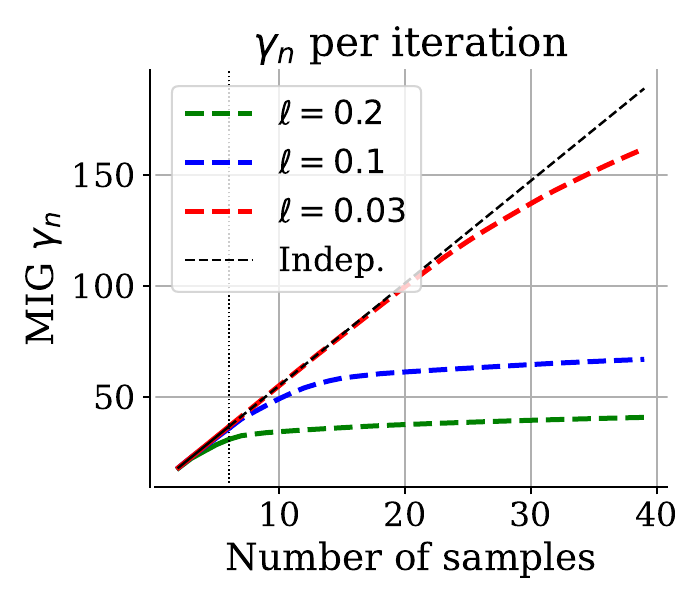}
\end{minipage}
    \caption{Three models (green, blue, red) with varying lengthscales, and thus varying complexity, attempting to model the same objective, acquiring data by greedily maximizing the IG. The MIG is shown for the three models as well as an independent kernel (dashed black), where the matrix $\mathbf{K} = \bm{I}$. The MIG for the complex model closely follows the independent kernel for 20 samples, suggesting that the complex model can acquire 20 data of approximately maximal variance. The vertical line in the MIG-plot indicates the current iteration.}
    \label{fig:compgp} 
\end{figure*}
\paragraph{A Working Definition of "Vanilla" BO\\}
In addition to the two main components — the probabilistic surrogate model and the acquisition function - BO entails multiple hidden design choices, that are paramount to its efficiency. We consider the vanilla BO algorithm to standardize the output values, and to use either a Squared Exponential (Radial Basis Function, RBF)~\cite{jones-jgo01a} or a $\frac{5}{2}$-Matérn~\cite{snoek-nips12a, snoek-icml14a} Kernel, with ARD lengthscales, an \ei{}-family acquisition function and multi-start gradient-based acquisition function optimization. 

While both MLE and MAP are commonly used for hyperparameter selection in practical Bayesian optimization, prevalent BO frameworks~\cite{balandat2020botorch, gpyopt2016, smac} employ MAP estimation, setting a prior on $p(\bm{\theta})$. While not included in our definition of the vanilla algorithm, the prior $p(\bm{\ell})$ commonly places high density on low values of $\bm{\ell}$. Furthermore, broad, uninformative priors are conventionally used on $\sigma^2_\varepsilon$ and $\sigma^2_f$. While fully Bayesian hyperparameter treatment~\cite{osborne2010bayesian, snoek-nips12a} may also be used, we do not consider it part of the vanilla algorithm.

\subsection{The Maximal Information Gain}\label{sec:mig}
Our work centers around the \textit{assumed} complexity of a problem, which conceptually could be seen as the size of the space of functions that have non-negligible probability under the GP prior. To quantify the assumed complexity, we use the \textit{Maximal Information Gain} (MIG)~\citep{srninivas-icml10a} measure, which is the maximum obtainable information about the function from querying a fixed number of points. Firstly, we recall the \textit{Information Gain} (IG) for a GP model and a set of points $\bm{X}$ is defined as 
\begin{equation}
    I(y_{\bm{X}}, f_{\bm{X}})  = \frac{1}{2} \log |\bm{I} + {\sigma_\varepsilon^{-2}} \mathbf{K}|,
\end{equation} 
where $\mathbf{K} = k(\bm{X}, \bm{X})$ is the Gram matrix for $\bm{X}$. Then, for a fixed number of points $|\setn| = n$, the MIG is the maximizer of this measure
\begin{equation}
\gamma_n = \max_{\setn\subset\mathcal{X}} I(y_\setn, f_\setn).
\end{equation}
For fixed observation noise, the MIG is fully defined by the covariance matrix, which in turn depends on the choice of kernel, the problem dimensionality and the kernel hyperparameters. The MIG is maximal when the samples are independent, i.e., $\mathbf{K} \approx \bm{I}$. The MIG lacks a closed form solution, but is approximated to $(1-1/e)$-accuracy by sequentially querying the set of points with maximal posterior variance~\cite{nemhauser1978submod, JMLR:v9:krause08a}.

Since the MIG measures the assumed complexity of $f$, it effectively quantifies the difficulty of optimizing a given task within a Bayesian optimization context~\cite{srninivas-icml10a, Srinivas_2012, bergenkamp2019noregret}, given that \emph{the assumptions on $k$ are accurate}. As long as the the MIG is nearly linear in the number of observations, there are regions in the search space that are almost independent of the collected data under the model. As such, there are still locations that we know nothing about, which makes optimization difficult. On the contrary, a small growth rate of the MIG suggests that the model would learn little by querying an additional point.

In Fig.~\ref{fig:compgp}, we provide an intuition for the MIG. We show a simpler model (left, green), as well as increasingly complex models (blue, red) for six data points. Their associated MIGs (right) are displayed for the current iteration (solid) and subsequent iterations (dashed) up until iteration 40. For the simpler model, there is little left to learn about the function, and as such, its subsequent MIG growth is small. \emph{If} the green model is accurate, subsequent optimization is trivial due to efficient modelling that stems from large correlation in the data. On the contrary, the almost-zero correlation displayed in the red model makes its optimization vastly more difficult. 
This point is further emphasized by its MIG, which starts to deviate substantially from an independent kernel first after 20 data points. This suggests that \emph{the model has capacity for 20 almost-maximal variance data acquisitions}, despite modelling only a one-dimensional objective.

\section{Related Work}\label{sec:vanillaborelated}
Multiple approaches have been proposed to tackle the limitations of BO in high dimensions. These resort to structural assumptions on the objective, which we outline by class.

\paragraph{Low-dimensional active subspaces\\} 
Subspace methods assume the existence of a lower-dimensional space which is representative of the function in the full-dimensional space. 
The active subspace can can be either axis-aligned~\citep{nayebi2019framework, pmlr-v161-eriksson21a, papenmeier2022increasing, papenmeier2023bounce} or non-axis-aligned~\citep{garnett2014active, wang2016bayesian, pmlr-v97-kirschner19a, binois2020random, letham2020re}. Explicit variable selection approaches~\cite{NIPS2013_8d34201a, li2017dropout, NEURIPS2020_e2ce14e8, song2022monte, hellsten2023highdimensional} employ the axis-aligned assumption to identify important variables to optimize over. 

\paragraph{Additive kernels\\} 
AddGP methods~\citep{duvenaud2011additive, kandasamy-icml15a, gardner-aistats17a, pmlr-v70-wang17h, pmlr-v84-rolland18a, Han_Arora_Scarlett_2021, pmlr-v202-ziomek23a} decompose the objective into a sum of low-dimensional component functions, where by assumption each component is only impacted by a small subset of all variables. As such, the maximal dimensionality of each component is substantially lower than the full dimensionality of $f$. 

\paragraph{Local Bayesian optimization\\} 
These approaches~\cite{eriksson2019turbo, GIBO, pmlr-v139-wan21b, NEURIPS2022_555479a2, wu2023the} adaptively restrict the search space to combat the CoD, limiting the optimization to a subset of the search space. By focusing on a smaller portion of the search space, the model exhibits less variation than a global model, which simplifies optimization. Moreover, enforcing local optimization decreases the susceptibility of the optimizer to the model.

\paragraph{Non-Euclidean kernels\\} 
These methods are employed to escape the exponential growth of the typical hypercube search space in the dimensionality of the problem. Cylindrical kernels ~\cite{swersky2014raiders, oh18bock} transform the geometry of the search space, which consequently expands the center of the search space, shrinking the boundaries.

The three pieces of related work that are most similar to ours are Elastic GPs~\cite{rana2017elastic},  SAASBO~\citep{pmlr-v161-eriksson21a} and BOCK~\cite{oh18bock}, which all perform optimization in the full-dimensional search space. Of these,~\cite{rana2017elastic} consider various lengthscales of the GP when optimizing the acquisition function, but uniquely does not impose simplifying assumptions on the model. \citep{pmlr-v161-eriksson21a} and BOCK~\cite{oh18bock} employ their aforementioned assumptions to facilitate effective optimization. Contrary to these works, we facilitate optimization in the ambient space without making any of the specific structural assumptions outlined in Sec.~\ref{sec:vanillaborelated}.

%

\section{Pitfalls of High-Complexity Assumptions}
\label{sec:issues}
We now discuss the issues related to highly complex models and connect it to the high-dimensional setting. Sec.~\ref{sec:companddim} demonstrates the intuitive relation between complexity and dimensionality. Building upon the intuitive understanding of the MIG and the related model design choices gained in Sec.~\ref{sec:mig}, we delve into the BO-specific pathology that arises from an overly complex model in Sec.~\ref{sec:boundary}, proving that it is distinct from the well-known \emph{boundary issue}~\cite{swersky2017improving}.
Thereafter, we demonstrate how various HDBO methods circumvent the high-complexity-issue by showing the  how conventional structural assumptions reduce the model complexity. In subsequent sections, we will use the terms MIG and complexity interchangeably. 

\subsection{Complexity and Dimensionality}\label{sec:companddim}
Increased model complexity most often becomes a critical issue for BO algorithms in high-dimensional problems — with increasing dimensions, the maximal space between points increases. Specifically, the expected distance between randomly sampled points in a unit cube increases proportional to the square root of the dimension~\cite{koppencurse}. 
For both the RBF and the Matern-$\frac{5}{2}$ kernels, this greatly impacts the covariance, which decreases exponentially with the $\bm{\ell}$-normalized squared distance.

\begin{figure}
    \centering
    \includegraphics[width=\linewidth]{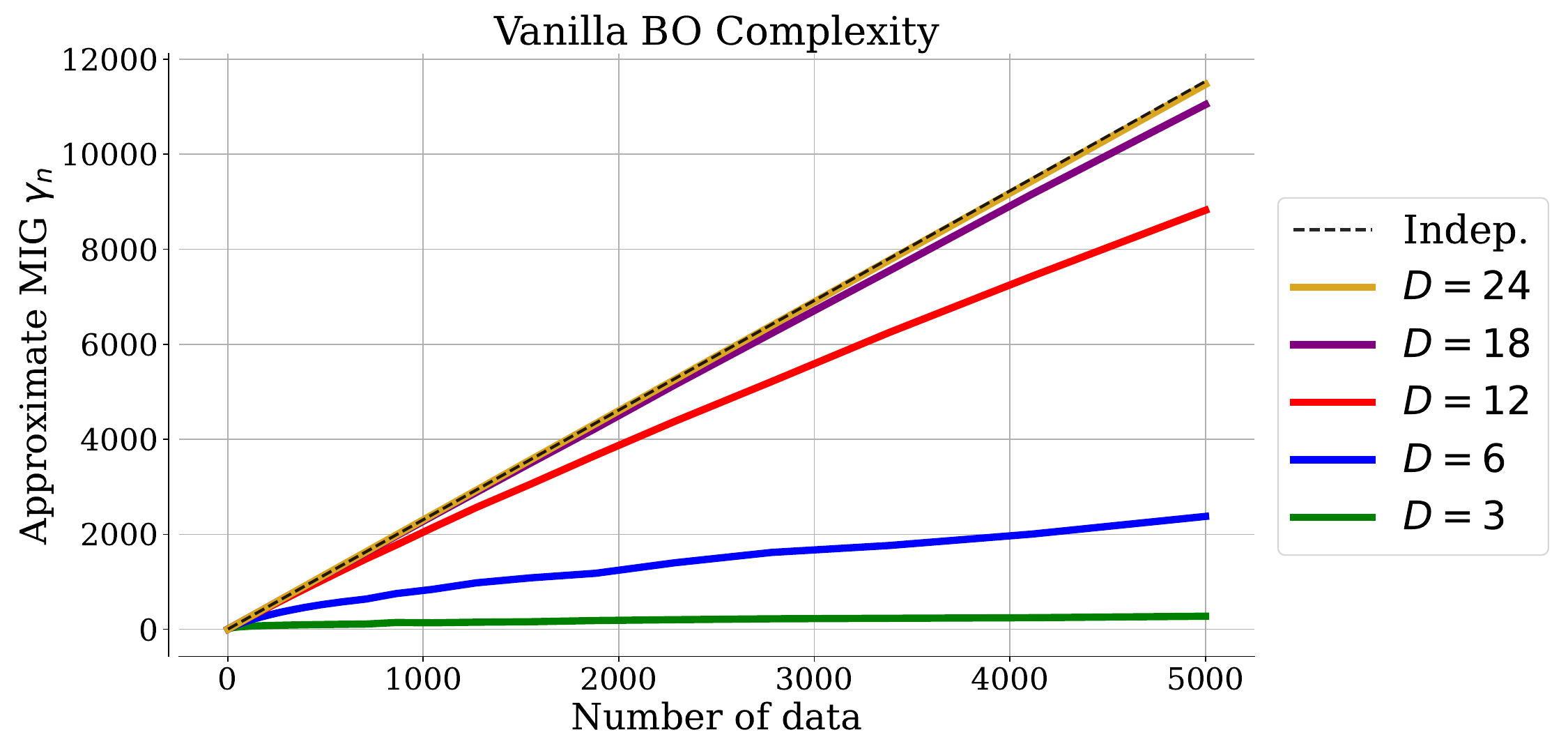}
    \caption{Complexity scaling in the number of data points for varying dimensionalities of the problem for vanilla BO with a lengthscale of $\bm{\ell} = 0.5$. For $D=18$, the complexity visually differs from an independent kernel after approximately 3000 data points. For $D=24$, 5000 data points are not sufficient to rid independence between observations. The MIG is approximated by sampling evenly distributed data using a SOBOL sequence.}
    \vspace{-3mm}
    \label{fig:hdvan}
\end{figure}

In Fig.~\ref{fig:hdvan}, we display the scaling of the complexity in the number of data points, considering an RBF-kernel with fixed lengthscales. The curves represent increasing dimensionalities.
As the dimensionality increases, the covariance matrix increasingly resembles that of an independent kernel (black dashed line). For $D=18$ (purple), this manifests in a visible difference first after 3000 samples, whereas for $D=24$ (yellow), 5000 samples is insufficient to distinguish an RBF kernel from an independent one, which implies that $k(\bm{X}_{5000}, \bm{X}_{5000})\approx \bm{I}$. This strongly suggests that global modelling of the objective is uninformative, as the model quickly reverts back to its prior mean and variance even after collecting vast amounts of data, and meaningful inference between observed data points becomes very difficult.

\subsection{The Boundary Issue Revisited}\label{sec:boundary}
As covered in Sec.~\ref{sec:mig}, high complexity implies that $k$ produces relatively low correlation both between acquired data points and in prospective queries. Thus, the GP will only be informative close to existing observations, and will quickly revert back to the prior as we move away from this data, as demonstrated in the rightmost model in Fig.~\ref{fig:compgp}. Under this regime, the BO data acquisition will be highly dependent on the hyperparameters that dictate said GP prior, namely the signal variance $\sigma_f^2$ and the mean constant $c$. We note that, while these parameters are not always learned~\cite{death2020mean}, the choice to fix them to 1.0 and 0.0 respectively influences on the behavior of the BO algorithm as well. 

Introduced by~\citet{swersky2017improving},  the boundary issue is the phenomenon that \ei{} will, in high-dimensional settings, repeatedly query uninformed, high-variance points along the boundary of the search space to maximally explore in light of an uninformed model. We contrast this claim by the following proposition, which demonstrates that \ei{} does \emph{not} tend towards maximal variance when the model is uninformed, namely when $\mathbf{K}\approx \bm{I}$ (as in the $D=24$ example in Fig.~\ref{fig:hdvan}). We denote by $\bm{x}_\text{inc}$ the location of the incumbent, its value by $y_\text{max}$, and the GP mean function by $c$.   

\begin{proposition}[\textbf{Lower Bound on EI Correlation}]$\quad$\label{th:boundary}
Assume that $y_{max} > c$, $\mathbf{K} = \sigma_f^2 \bm{I}$ and that the candidate query $\bm{x}_*$ correlates with at most one observation. Then, the correlation $\rho^* = \sigma_f^{-2}k(\bm{x}_*, \bm{x}_{inc})$ between the next query $\bm{x}_* =\argmax_{\bm{x}\in \mathcal{X}} \ei{}(\bm{x})$ and $\bm{x}_{inc}$ satisfies 
\begin{equation}\label{eq:bound}
    \rho^*\sqrt{\frac{1+\rho^*}{1-\rho^*}} \geq \frac{y_{max} - c}{\sigma_f}.
\end{equation}

\end{proposition}
We outline the proof in App.~\ref{app:boundary}, where we parameterize \ei{} by the correlation $\rho$ between a candidate $\bm{x}_*$ and the incumbent, and show that $\frac{\partial\ei{}}{\partial\rho}$ is positive for all values of $\rho$ below the bound in Eq.~\ref{eq:bound}. Thus, \ei{} will prefer an observation that has substantial correlation with $\bm{x}_{inc}$ to one that does not.

Proposition~\ref{th:boundary} demonstrates that, when correlation in the model is low, \ei{} does not seek out high-variance regions as described by ~\citet{swersky2017improving}. This  contrasts the common belief that HDBO intrinsically suffers from excessive exploration around the borders~\cite{Siivola2017CORRECTINGBO, oh18bock, eriksson2019turbo, pmlr-v161-eriksson21a, GIBO, binois2022hdsurvey, Eduardo2022BayesianOW}. As we will soon cover, uninformative models frequently display the opposite behavior, where repeated queries are made exceedingly close to the incumbent.

\begin{figure}[tb]
    \centering
\begin{minipage}[b]{0.585\linewidth}
    \includegraphics[width=\textwidth]{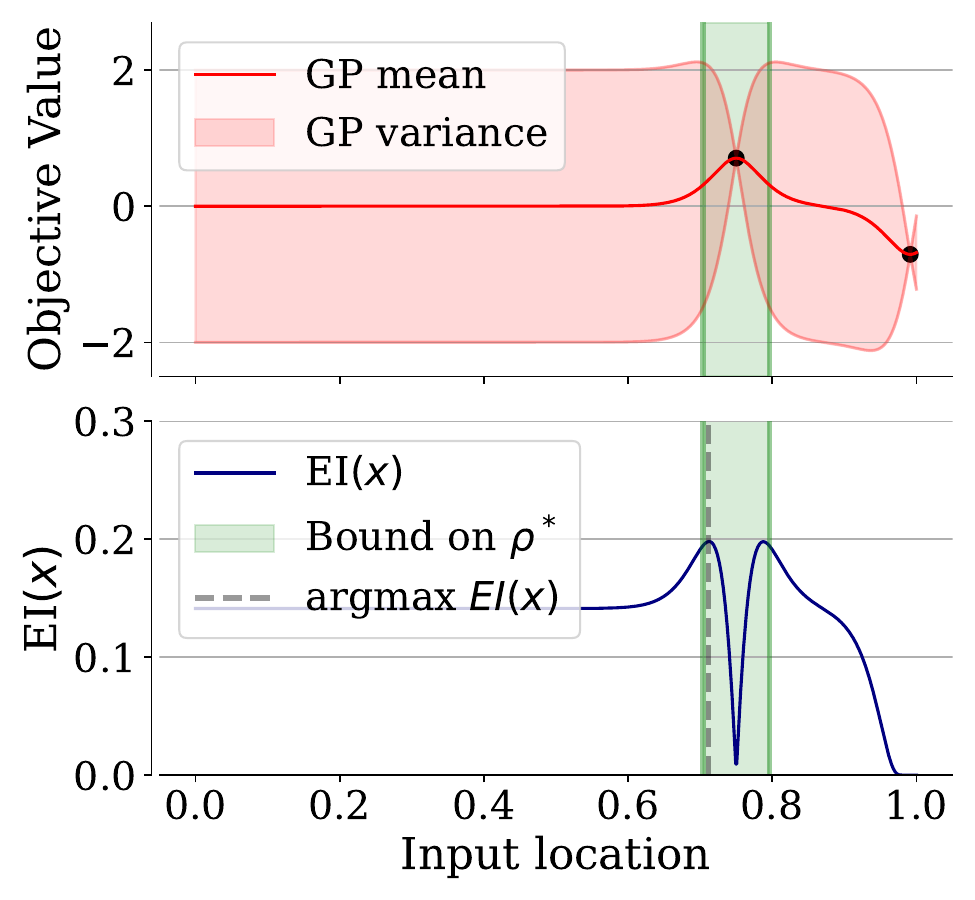}
    \vspace{-2mm}
    \textbf{a)}
    \vspace{-1mm}
\end{minipage}
\hfill
\begin{minipage}[b]{0.39\linewidth}
    \includegraphics[width=\linewidth]{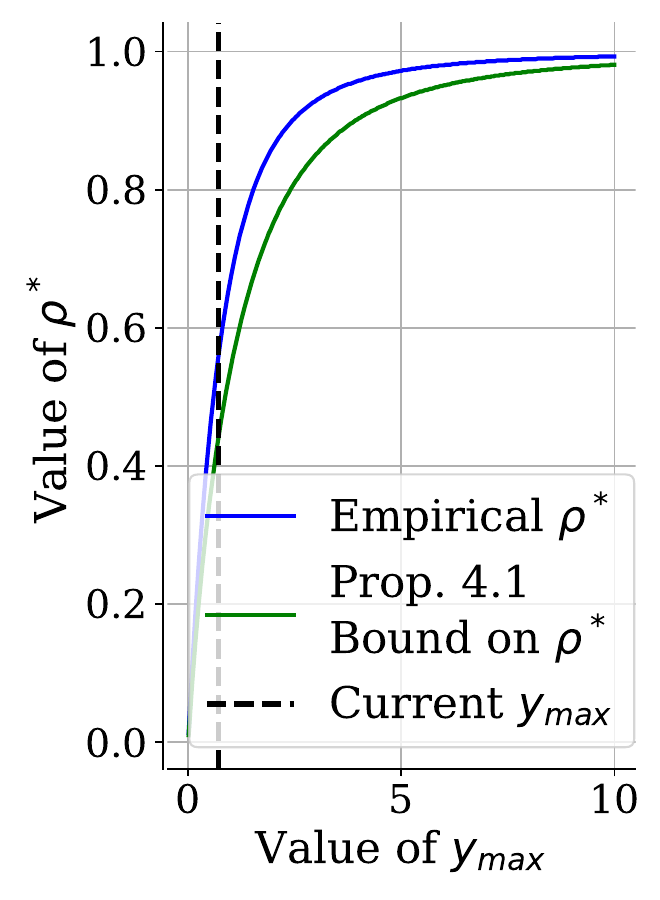}
        \vspace{-2mm}
    \textbf{b)}
    \vspace{-1mm}
\end{minipage}
    \caption{Lower bound on the optimal correlation $\rho^*$ between the incumbent and the upcoming query. \textbf{a)} The GP for two almost-independent observations with a large exploratory region. \ei{} prefers to query close to the incumbent, well within the bound on $\rho^*$ from Prop.~\ref{th:boundary}. \textbf{b)} Tightness of the bound compared a numerical solve for optimal correlation for various values of $y_{max}$.}
    \label{fig:bound}
\end{figure}

In Fig.~\ref{fig:bound}a, we observe this query behavior in action for the high-complexity model in Fig.~\ref{fig:compgp} —  despite many large-variance regions, the next query is very close to the current best, and within our correlation bound. 
In Fig.~\ref{fig:bound}b, we display the numerical solution to $\frac{\partial\ei{}}{\partial\rho} = 0$ together with the analytical bound in Proposition~\ref{th:boundary}.
We observe that for typical values of the GP mean and outputscale, the candidate query has substantial correlation with the incumbent under the aforementioned setting of an uninformed model.

With that said, the phenomenon of frequent querying of the boundary may still occur in practice, which we expand on in App.~\ref{app:boundary_emp}. Specifically, querying of the boundary may occur when lengthscales are very long along one or more dimensions, which can occur when fitting the model using MLE~\cite{NIPS1995_7cce53cf, rasmussen-book06a, JMLR:v24:22-1153}. Then, candidates will be low-variance and highly correlated with existing data, despite being located at the boundary of one or multiple dimensions that are all deemed irrelevant. In Fig.~\ref{fig:mle_locality} visualizes this setting for a 2D toy example. 




Proposition~\ref{th:boundary} establishes that \ei{} does not intrinsically pursue high-variance, uninformed regions. Rather, queries preferentially have substantial correlation with the incumbent. In a high-complexity setting, substantial correlation only arises when a data point is close to existing data, which suggests that \ei{} should make queries in close proximity to the incumbent. 
As a result, the algorithm tends to very seldom query far from the incumbent, resulting in an exploitative behavior with similar qualities to local search.
This phenomenon arises when there is negligible correlation in the data, namely when model complexity is too high to effectively model the objective function with existing data. In App.~\ref{app:boundary_emp}, we expand on the local search behavior of the EI acquisition function, demonstrating its prevalence in conventional setups for dimensionalities as low as six.


\subsection{Complexity of Existing HDBO}
\begin{figure}
    \centering
    \includegraphics[width=\linewidth]{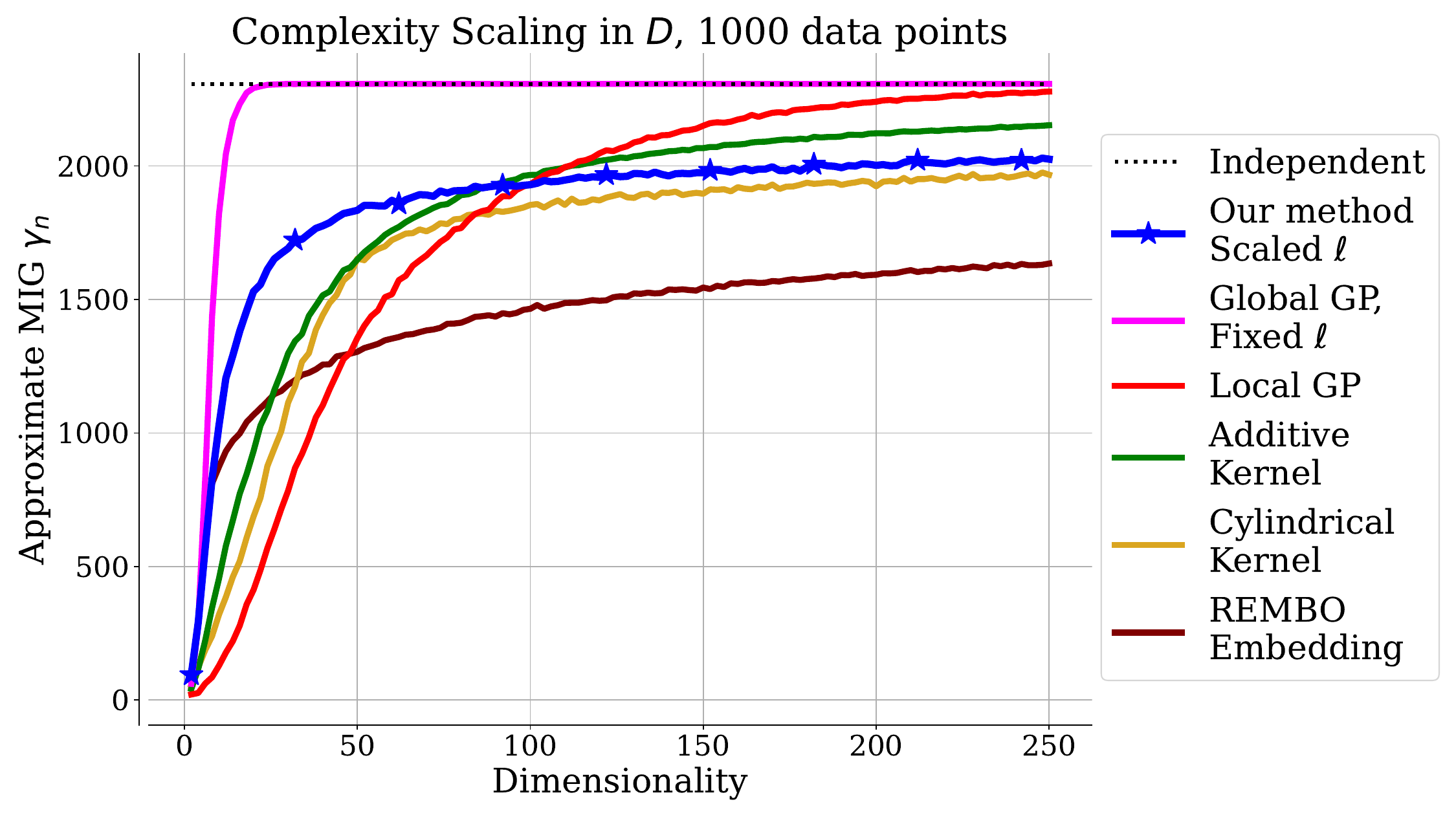}
    \caption{We display the model complexity scaling in the dimensionality of the problem for 1000 data points for various HDBO algorithms. Vanilla BO with fixed lengthscales (magenta) approaches independent complexity at approximately 20 dimensions. As expected, REMBO random embeddings (brown) reduce complexity the most, followed by BOCK cylindrical kernels (yellow). The MIG growth of our proposed modification of the global GP (blue) flattens out at a rate similar to cylindrical kernels (yellow), despite modelling the original, full-dimensional space.}
    \label{fig:hdcomp}
\end{figure}

Having established that high complexity can yield uninformative models, as well as having discussed the link between complexity and dimensionality, it is evident that complexity assumptions must be sensible to facilitate a calibrated GP in HDBO. Notably, all the classes of HDBO algorithms outlined in Sec.~\ref{sec:vanillaborelated} have such complexity-lowering assumptions. In Fig.~\ref{fig:hdcomp}, we display the modeled complexity of the most common classes of HDBO algorithms. 
Fig.~\ref{fig:hdcomp} can be viewed as a cross section of Fig.~\ref{fig:hdvan}, where we fix the number of data points to 1000 and instead vary the dimensionality of the problem to demonstrate how each HDBO class lower the growth of the complexity in the dimensionality, relative to a common global GP model as well as an independent kernel. The methods presented are: REMBO~\cite{wang2016bayesian} with $d_e=4$, random AddGPs~\cite{pmlr-v202-ziomek23a}, BOCK~\cite{oh18bock}, Local GPs~\cite{eriksson2019turbo} after one round of shrinkage, the global GP with fixed lengthscales from Fig.~\ref{fig:hdvan}, and our proposed method — scaling the lengthscales in the dimensionality of the problem, which is introduced later in Sec.~\ref{sec:method}. While each algorithm has parameters that affect the MIG, we have set parameters to make the comparison as fair as possible. In App.~\ref{app:migcomp}, we outline the setup in detail.

As we have observed previously, employing a full-dimensional GP without restrictions is far too complex, as we have approximate independence after 1000 observations already for an $18D$-objective. As expected, random subspace methods have small complexity increase in the ambient dimensionality. The increase stems from the fact that random, non-axis-aligned embeddings may slice the ambient dimensions very narrowly, which results in shorter lengthscales on the embedded model. The complexity increase for both our method (blue) and cylindrical kernels (yellow) stagnates rapidly, effectively assuming only marginal complexity increases after $D=100$. Local methods (red) scale at same rates as global GPs (blue), but work on a drastically simplified model due to the lengthscale-scaled trust regions.

We re-iterate that low complexity is not strictly a desirable property, but as per Occam's razor, the most desirable property is to have the lowest possible complexity for a model that sufficiently aligns with the objective. This is especially true in the context of small data optimization, where each new data point acquired and employed to train the model is costly. We note, however, that the almost-independence exhibited by the global GP in Fig.~\ref{fig:hdcomp} (magenta) for even moderate dimensionalities inevitably leads to the degeneracy highlighted in Sec.~\ref{sec:boundary}. Moreover, the assumptions behind each HDBO method are all means to the same end - reducing model complexity to manageable levels.

\section{Low-complexity High-dimensional\\Bayesian Optimization}
\label{sec:method}
\noindent
Hypothesizing that the pitfalls of HDBO are strictly caused by assumptions of insurmountable complexity, we present our main methodological contribution. We design a simple, plug-and-play assumption that retains almost constant complexity as the dimensionality increases. Similar to Fig.~\ref{fig:compgp}, we achieve this by adjusting the prior on the lengthscales to the dimensionality of the problem to the task at hand. Moreover, we ensure a calibrated signal variance by drawing on previous findings on GPs in an over-parameterized regime.

\subsection{Ensuring Meaningful Correlation}
Since stationary kernels compute covariances based on distances between data and both the diagonal and the distance between randomly sampled points in a $D$-dimensional hypercube scales as $\sqrt{D}$~\cite{koppencurse}, increasing the lengthscales at this rate, $\ell_i \propto \sqrt{D}$, counteracts the complexity increase that stems from the increased distances.
This change may, for example, be achieved by scaling the $\mu$ term of a LogNormal ($\mathcal{LN}$) prior
\begin{equation}
    \ell_i \sim \mathcal{LN}\left(\mu_0 + \frac{\log(D)}{2}, \sigma_0\right)
\end{equation}

where $(\mu_0, \sigma_0)$ are suitable parameters of $p(\bm{\ell})$ for a one-dimensional objective. This shifts both the mode and mean of the distribution by a factor of $\sqrt{D}$. Notably, our method does not increase the number of hyperparameters in a MAP-based BO setup. Furthermore, the proposed change may similarly be applied the more commonly used Gamma prior, with a different parameterization~\cite{2004JApMe..43.1586C}. Importantly, the change in complexity is not definitive, as we may still find some variables to be more important than others and adjust on-the-fly through MAP estimation of $\bm{\ell}$.

The proposed change suggests that the problem of large distances between points, and thereby the insurmountable complexity, is one that arises by assumption. Specifically, the lengthscale priors $p(\bm{\ell})$ employed by conventional BO frameworks~\cite{balandat2020botorch, spearmint, smac} place substantial density on low values of $\bm{\ell}$. Assuming that all dimensions are of major importance may appear like a conservative and sensible choice. For moderately high dimensions, however, it practically guarantees that the problem will be impossible to model globally, even with the largest of budgets. Our method takes the opposite approach, and simply assumes that a problem \textit{is simple enough to be modeled globally, for any dimensionality.}
\subsection{Calibrating Epistemic Uncertainty}
Lastly, we consider the role of the signal variance parameter, whose impact on data acquisition is evidenced by Prop.~\ref{th:boundary}. Motivated by findings on on the optimal value $\hat{\sigma}_f^2$ of $\sigma_f^2$ generally in~\citet{Moore_2016} and in the over-parameterized regime by~\citet{pmlr-v161-ober21a}, we consider
\begin{equation}
    \hat{\sigma}_f^2 = \frac{1}{n} \bm{y} ^\text{T}\mathbf{K}^{-1}\bm{y}
\end{equation}

\begin{figure*}[h!]
    \centering
    \includegraphics[width=0.82\linewidth]{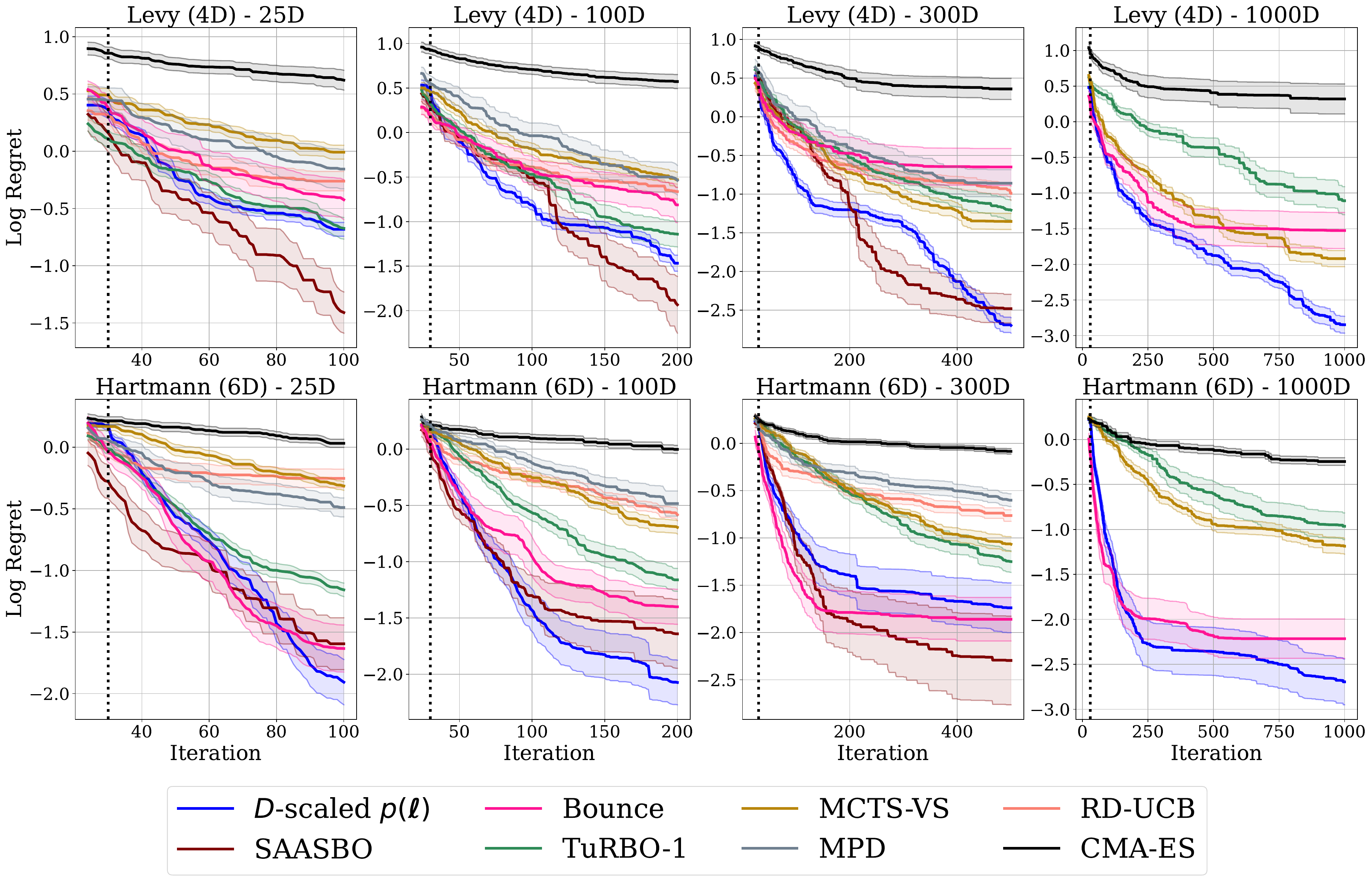}
    \caption{Average log regret of all baselines on Levy (4D) and Hartmann (6D) 
    synthetic test functions of varying ambient dimensionality across 20 repetitions (10 for SAASBO). Vanilla BO performs second best, beaten only by SAASBO on four tasks, whose axis-aligned subspace assumption (along with MCTS-VS' variable selection) aligns perfectly with the task at hand. We omit SAASBO from the 1000D benchmarks due to the prohibitive runtime, and RD-UCB and MPD due to a combination of runtime and numerical instability.}
    \label{fig:synthetic}
\end{figure*}

which, in a BO context, has a different impact than in GPs generally. As we are able to selectively acquire our data, a large number of parameters and substantially correlated data will simplify data fit, driving down the optimal value of $\sigma_f^2$.  When data is repeatedly re-normalized, the issue will be reinforced, as the signal variance is further decreased, and another highly correlated query is selected. As such, we fix $\sigma_f^2 = 1$ to match the scale of the standardized observations, and to ensure that $\sigma_f^2$ does not diminish over time. In App.~\ref{app:opslearn}, we ablate this modeling choice and demonstrate its impact on the value of the outputscale during optimization for two real-world optimization tasks, and in App. \ref{sec:opsperf}, we demonstrate the impact of the outputscale modeling choice on optimization performance.



\section{Results}\label{sec:results}
We now compare our Vanilla Bayesian Optimization method with a dimensionality-scaled lengthscale prior (we will refer to our method as $D$-scaled $p(\bm{\ell)}$ or DSP for clarity), against state-of-the-art HDBO methods. We will include various classes of HDBO methods, such as the subspace-methods Bounce and SAASBO~\cite{papenmeier2023bounce,pmlr-v161-eriksson21a}, the Local BO algorithms TuRBO~\cite{eriksson2019turbo} and Maximal Probability of Descent~\cite{NEURIPS2022_555479a2} (MPD), the AddGP method RD-UCB~\citep{pmlr-v202-ziomek23a}, the variable selection method MCTS-VS~\cite{song2022monte}, and CMA-ES~\cite{hansen-eda06a}. We use each method's official repository, with the exception of SAASBO which is run through Ax~\cite{bakshy2018ae}. We detail the experimental setup in App.~\ref{app:expsetup}, and our code is publicly available at~\url{https://github.com/hvarfner/vanilla_bo_in_highdim}.

We instantiate the DSP with $\mu_0 = \sqrt{2}, \sigma_0 = \sqrt{3}$, which equates to $\bm{\ell} \approx 0.50$ for $D=6$ under the mode of $p(\bm{\ell})$. We initialize all methods with 30 samples, marked by a dashed vertical line. Bounce, CMA-ES and MPD deviate from conventional initialization.  On all benchmarks, we use \logei{}~\cite{ament2023unexpected}, using a low acquisition optimization budget of 512 initial (global) SOBOL samples and 512 Gaussian samples around the incumbent, followed by L-BFGS on the 4 best candidates, which is made possible by the low-complexity-high-smoothness model. In App.~\ref{app:abl}, we demonstrate that our method delivers consistent performance for a range of complexities.
 
\begin{figure*}[h!]
    \centering
    \includegraphics[width=0.95\linewidth]{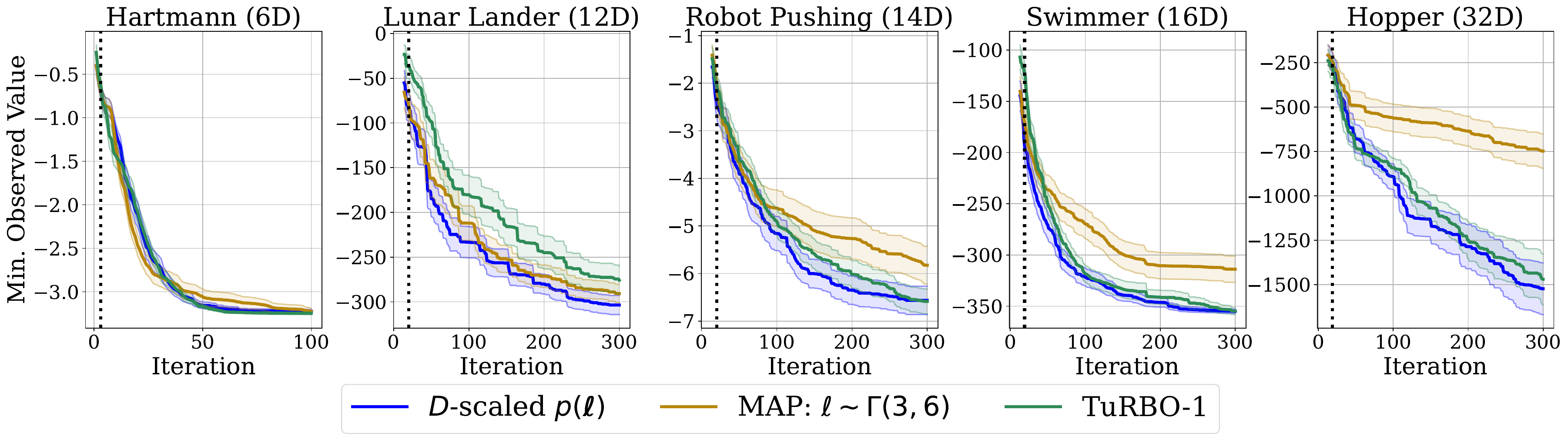}
    \vspace{-1mm}
    \caption{Best observed value for the DSP, conventional MAP, and TuRBO on Hartmann (6D) and four mid-dimensional real-world optimization tasks. All methods perform comparably on Hartmann, while the DSP outperforms or is on part with TuRBO on the other tasks. Notably, the DSP performs at least equally well as the $\Gamma(3, 6)$ on all tasks, and substantially better on 4 out of 5 tasks, which suggests that it is well-suited as a drop-in replacement for conventional priors.}
    \label{fig:mid}
    \vspace{-1mm}
\end{figure*}
\subsection{Sparse Synthetic Test Functions}
We start by evaluating the DSP on a collection of commonly considered synthetic test functions with varying \textit{total} and \textit{effective dimensionality}. We note that the assumptions made in Sec.~\ref{sec:method} diametrically oppose these test cases - each function has a low number of highly important dimensions with the large remainder being unimportant, whereas we assume that \textit{each} dimension has relatively small impact. 

The DSP is highly performant, as it rapidly identifies the important dimensions and subsequently optimizes the task. This is similar in  to~\cite{pmlr-v161-eriksson21a}, whose assumptions, represented through its sparse lengthscale prior, aligns perfectly with the task at hand. As such, SAASBO should, and does, perform best on average, with Vanilla BO being second. Notably, as Vanilla BO does not require the HMC~\cite{bingham2018pyro} fully Bayesian model fitting that SAASBO uses, it runs in a small fraction of the time.

\subsection{A Plug-in on Mid-Dimensional Tasks}
Subsequently, we use the DSP as a plug-in for low- and mid-dimensional tasks, primarily those considered in~\citep{eriksson2019turbo}, to evaluate its ability to serve as a substitute for conventional, non-adaptive hyperparameter priors. The Lunar Lander (12D) and Robot Pushing (14D) tasks from~\citep{pmlr-v70-wang17h}, as well as the Swimmer (16D) and Hopper (32D) reinforcement learning tasks from the MuJoCo suite~\cite{todorov2012mujoco}, where we aim to learn a linear policy for two objects with varying degrees of freedom. We evaluate against a $\Gamma(3, 6)$ lengthscale prior with learnable $\sigma^2_f$, and against TuRBO, commonly considered the state-of-the-art mid-dimensional BO method. In Fig.~\ref{fig:mid}, it is shown that the DSP is either competitive with, or outperforms, TuRBO on all tasks. In App.~\ref{app:boundary}, we empirically display
the pronounced local search-like behavior of the $\Gamma(3, 6)$ prior for Hartmann (6D), which suggests that it does not offer a calibrated exploration-exploitation trade-off. On the contrary, the DSP maintains a moderate distance between queries, which indicates a calibrated trade-off throughout optimization.

\subsection{High-dimensional Optimization Tasks}\label{sec:realworld}
\begin{figure*}[h!]
    \centering
    \includegraphics[width=\linewidth]{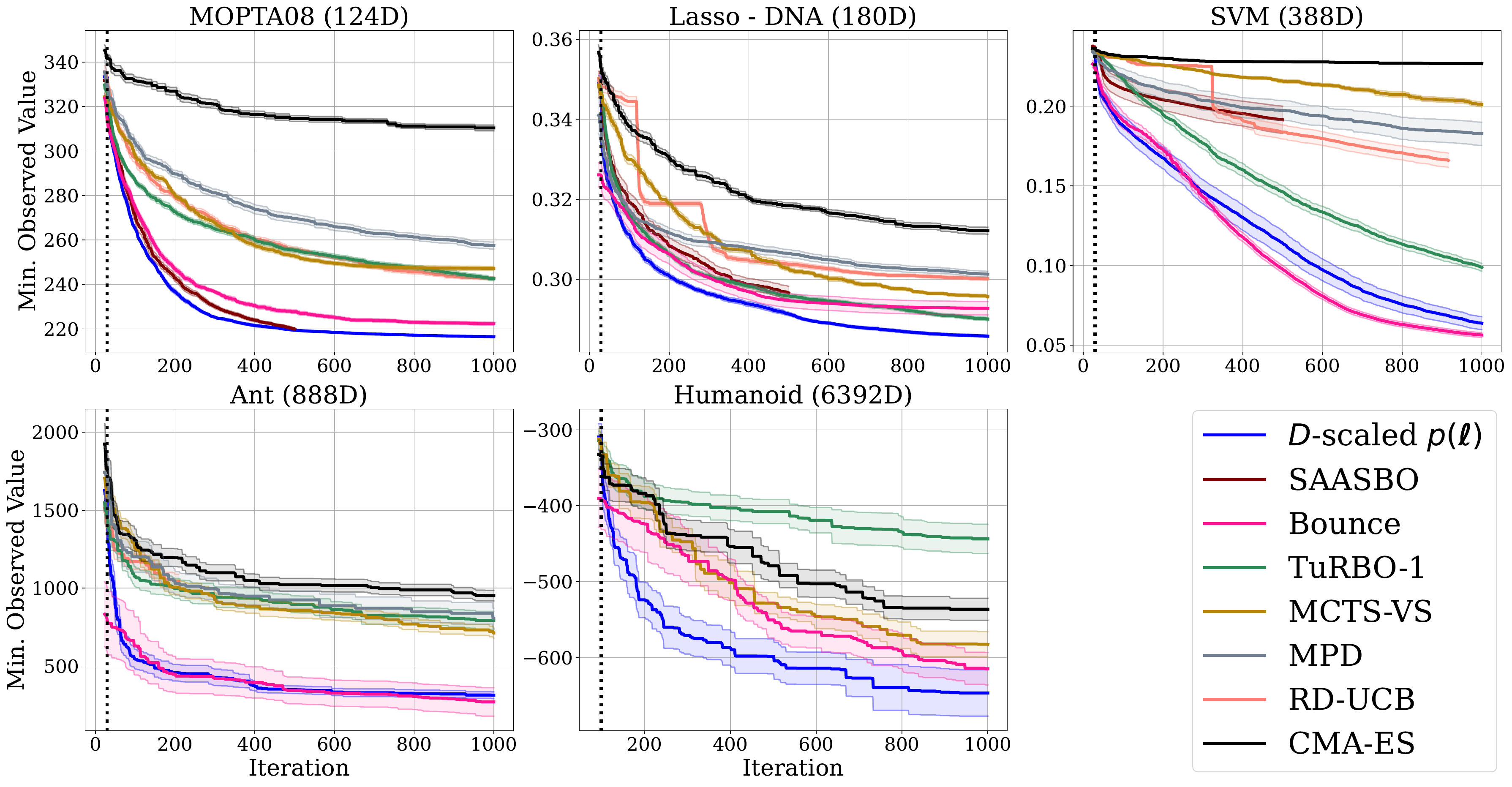}
    \caption{Best observed value of all baselines on five real-world tasks from of various domains across 20 repetitions (10 for SAASBO). Vanilla BO performs best across all tasks except for MOPTA, where it is outperformed by SAASBO, and the MuJoCo Ant, where BAxUS gets an advantage from performing the initialization phase in a lower-dimensional subspace, which enables it to obtain initial samples close to the center of the search space. Vanilla BO substantially outperforms all baselines on Lasso-DNA, SVM and Humanoid. Notably, the extreme dimensionality of Humanoid does not have an apparent negative impact the performance of Vanilla BO.}
    \label{fig:real}
\end{figure*}
We now benchmark Vanilla BO with the DSP against a collection of frequently considered tasks in the high-dimensional literature~\cite{pmlr-v161-eriksson21a, eriksson2019turbo, papenmeier2022increasing, papenmeier2023bounce, vsehic2021lassobench}:
Specifically, we consider MOPTA08 (124D), SVM (388D), Lasso-DNA (180D), and the MuJoCo~\cite{todorov2012mujoco} Ant (888D) and Humanoid (6392D) reinforcement learning tasks. We stress that, for all benchmarks where applicable, (BAxUS on SVM, Lasso-DNA and Humanoid, TuRBO on MOPTA and SVM, RD-UCB on Lasso-DNA), baselines perform within the error bars of the original implementation~\cite{papenmeier2022increasing, pmlr-v202-ziomek23a} or in other papers by the same authors in the case of TuRBO~\cite{eriksson2019turbo, pmlr-v161-eriksson21a}. 
In App.~\ref{app:anytime}, we display the performance in a lower-budget setting. In App.~\ref{app:mapmle}, we compare to conventional MAP and MLE.
\begin{figure}[h!]
    \centering
    \includegraphics[width=0.95\linewidth]{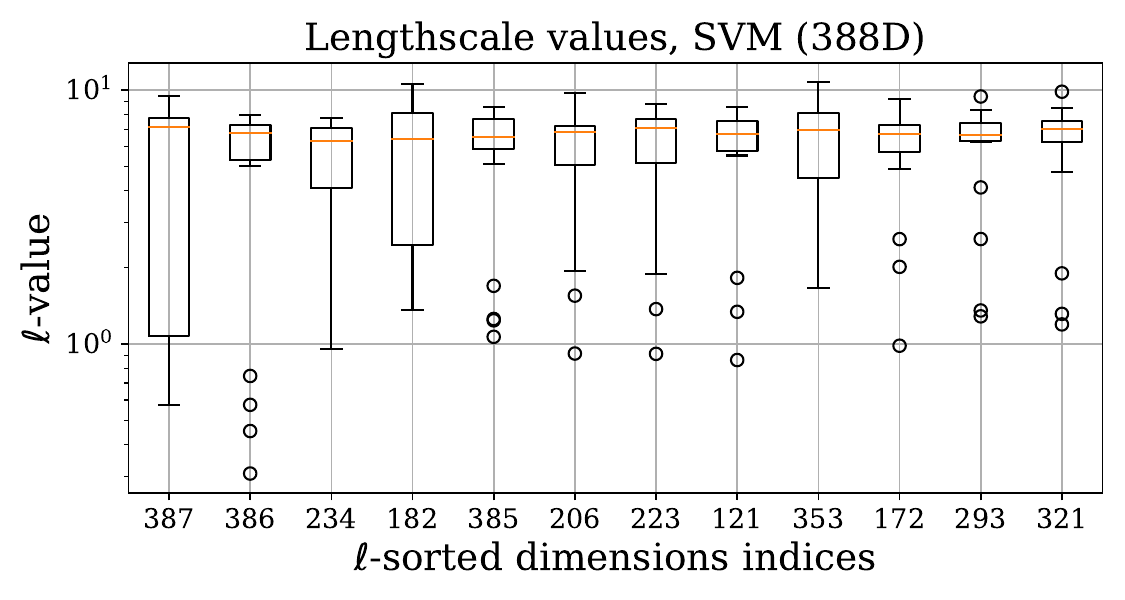}
    \caption{Distribution of lengthscale values for the DSP on the 388D SVM task. Lengthscales are sorted according to their mean log value. The three last indexed dimensions (385, 386, 387) are considered particularly active~\cite{pmlr-v161-eriksson21a}, but are not identified as such consistently in our method. the black horizontal lines indicate upper and lower quartiles, and the orange horizontal lines indicate the median.}
    \label{fig:svmhps}
\end{figure}
Fig.~\ref{fig:real} shows that Vanilla BO with the DSP is highly competitive, performing the best by a substantial margin on Lasso-DNA and Humanoid, and produces top-two performance on the remaining tasks. On the MuJuCo Ant, Bounce's low-dimensional initialization allows it to obtain an average value of 800 after DoE due to consistently sampling data points close to the center of the search space. Notably, the DSP is very consistent between repetitions, as evident by the small error bars. This can be attributed to the consistent modelling, as the DSP is not dependent on randomness in subspace design, trust region initialization or variable selection. Rather, it obtains a consistent, calibrated model through meaningful correlation in the data, from which it can effectively infer promising regions and improve upon the DoE. 

Notably, the DSP does \textit{not} heavily depend on identification of active variables. In Fig.~\ref{fig:svmhps}, we demonstrate for the distribution of lengthscale values for the 388D SVM after 250 iterations. The DSP does not consistently identify active dimensions with large confidence. Instead, the calibrated complexity of the model allows for meaningful inference along all dimensions, until particularly active dimensions are potentially identified. As such, the DSP does not \textit{require} the identification of active variables to achieve calibrated BO, but its identification helps optimization. As such, we attribute the superior performance of our method to the calibrated complexity, and the effective inference and exploration-exploitation trade-off that stems from it. 

\section{Conclusion and Future Work}
The curse of dimensionality has long been assumed to hinder the application of conventional Bayesian optimization in high dimensions. We show that the hindrance is not driven by dimensionality, but by the assumed complexity of the objective. We make minor modifications to the assumptions of the vanilla BO algorithm to make complexity scaling manageable with increasing dimensionality. As a result, we demonstrate that vanilla BO is extremely effective for problems of high dimensionality, outperforming the state-of-the-art for problems with dimensions into the thousands.

Nevertheless, we do not believe that tailored high-dimensional BO algorithms are unwarranted: if the problem at hand is known to adhere to the structural assumptions that are conventionally made (effective subspace, additivity) or where non-stationarity in the objective facilitates local modelling, we believe these approaches will be superior to the vanilla algorithm. However, these restrictive assumptions should not be made \textit{out of necessity}, but when prior knowledge supports them. For future work, we plan to investigate the topic of complexity as it relates to modelling in Bayesian optimization more broadly, and in the context of latent space GP models~\cite{Griffiths2017ConstrainedBO, NEURIPS2022_ded98d28}.
\newpage
\section*{Impact Statement}
This paper presents work whose goal is to advance the field of Machine Learning. There are many potential societal consequences of our work, none which we feel must be specifically highlighted here.

\section*{Acknowledgements}
Carl Hvarfner, Erik Hellsten and Luigi Nardi were partially supported by the Wallenberg AI, Autonomous Systems and Software Program (WASP) funded by the Knut and Alice Wallenberg Foundation. Luigi Nardi was partially supported by the Wallenberg Launch Pad (WALP) grant nbr. 2021.0348. The computations and data handling were enabled by resources provided by the National Academic Infrastructure for Supercomputing in Sweden (NAISS), partially funded by the Swedish Research Council through grant agreement no. 2024/5-228 and 2023/22-1024.
\bibliographystyle{icml2024}
\bibliography{bibliography/local,bibliography/lib,bibliography/proc,bibliography/strings}

\begin{thebibliography}{71}
\providecommand{\natexlab}[1]{#1}
\providecommand{\url}[1]{\texttt{#1}}
\expandafter\ifx\csname urlstyle\endcsname\relax
  \providecommand{\doi}[1]{doi: #1}\else
  \providecommand{\doi}{doi: \begingroup \urlstyle{rm}\Url}\fi

\bibitem[Ament et~al.(2023)Ament, Daulton, Eriksson, Balandat, and Bakshy]{ament2023unexpected}
Ament, S., Daulton, S., Eriksson, D., Balandat, M., and Bakshy, E.
\newblock Unexpected improvements to expected improvement for bayesian optimization.
\newblock In \emph{Thirty-seventh Conference on Neural Information Processing Systems}, 2023.
\newblock URL \url{https://openreview.net/forum?id=1vyAG6j9PE}.

\bibitem[Bakshy et~al.()Bakshy, Dworkin, Karrer, Kashin, Letham, Murthy, and Singh]{bakshy2018ae}
Bakshy, E., Dworkin, L., Karrer, B., Kashin, K., Letham, B., Murthy, A., and Singh, S.
\newblock Ae: A domain-agnostic platform for adaptive experimentation.

\bibitem[Balandat et~al.(2020)Balandat, Karrer, Jiang, Daulton, Letham, Wilson, and Bakshy]{balandat2020botorch}
Balandat, M., Karrer, B., Jiang, D.~R., Daulton, S., Letham, B., Wilson, A.~G., and Bakshy, E.
\newblock Botorch: A framework for efficient monte-carlo bayesian optimization.
\newblock In \emph{Advances in Neural Information Processing Systems}, 2020.
\newblock URL \url{http://arxiv.org/abs/1910.06403}.

\bibitem[Baptista \& Poloczek(2018)Baptista and Poloczek]{pmlr-v80-baptista18a}
Baptista, R. and Poloczek, M.
\newblock {B}ayesian optimization of combinatorial structures.
\newblock In Dy, J. and Krause, A. (eds.), \emph{Proceedings of the 35th International Conference on Machine Learning}, volume~80 of \emph{Proceedings of Machine Learning Research}, pp.\  462--471. PMLR, 10--15 Jul 2018.
\newblock URL \url{https://proceedings.mlr.press/v80/baptista18a.html}.

\bibitem[Berkenkamp et~al.(2019)Berkenkamp, Schoellig, and Krause]{bergenkamp2019noregret}
Berkenkamp, F., Schoellig, A.~P., and Krause, A.
\newblock No-regret bayesian optimization with unknown hyperparameters.
\newblock \emph{Journal of Machine Learning Research}, 20\penalty0 (50):\penalty0 1--24, 2019.
\newblock URL \url{http://jmlr.org/papers/v20/18-213.html}.

\bibitem[Bingham et~al.(2018)Bingham, Chen, Jankowiak, Obermeyer, Pradhan, Karaletsos, Singh, Szerlip, Horsfall, and Goodman]{bingham2018pyro}
Bingham, E., Chen, J.~P., Jankowiak, M., Obermeyer, F., Pradhan, N., Karaletsos, T., Singh, R., Szerlip, P., Horsfall, P., and Goodman, N.~D.
\newblock {Pyro: Deep Universal Probabilistic Programming}.
\newblock \emph{Journal of Machine Learning Research}, 2018.

\bibitem[Binois \& Wycoff(2022)Binois and Wycoff]{binois2022hdsurvey}
Binois, M. and Wycoff, N.
\newblock A survey on high-dimensional gaussian process modeling with application to bayesian optimization.
\newblock \emph{ACM Trans. Evol. Learn. Optim.}, 2\penalty0 (2), aug 2022.
\newblock \doi{10.1145/3545611}.
\newblock URL \url{https://doi.org/10.1145/3545611}.

\bibitem[Binois et~al.(2020)Binois, Ginsbourger, and Roustant]{binois2020random}
Binois, M., Ginsbourger, D., and Roustant, O.
\newblock On the choice of the low-dimensional domain for global optimization via random embeddings.
\newblock \emph{J. of Global Optimization}, 76\penalty0 (1):\penalty0 69–90, jan 2020.
\newblock ISSN 0925-5001.
\newblock \doi{10.1007/s10898-019-00839-1}.
\newblock URL \url{https://doi.org/10.1007/s10898-019-00839-1}.

\bibitem[Bull(2011)]{bull-jmlr11a}
Bull, A.~D.
\newblock Convergence rates of efficient global optimization algorithms.
\newblock 12:\penalty0 2879--2904, 2011.

\bibitem[{Cho} et~al.(2004){Cho}, {Bowman}, and {North}]{2004JApMe..43.1586C}
{Cho}, H.-K., {Bowman}, K.~P., and {North}, G.~R.
\newblock {A Comparison of Gamma and Lognormal Distributions for Characterizing Satellite Rain Rates from the Tropical Rainfall Measuring Mission.}
\newblock \emph{Journal of Applied Meteorology}, 43\penalty0 (11):\penalty0 1586--1597, November 2004.
\newblock \doi{10.1175/JAM2165.1}.

\bibitem[De~Ath et~al.(2020)De~Ath, Fieldsend, and Everson]{death2020mean}
De~Ath, G., Fieldsend, J.~E., and Everson, R.~M.
\newblock What do you mean? the role of the mean function in bayesian optimisation.
\newblock In \emph{Proceedings of the 2020 Genetic and Evolutionary Computation Conference Companion}, GECCO '20, pp.\  1623–1631, New York, NY, USA, 2020. Association for Computing Machinery.
\newblock ISBN 9781450371278.
\newblock \doi{10.1145/3377929.3398118}.
\newblock URL \url{https://doi.org/10.1145/3377929.3398118}.

\bibitem[Djolonga et~al.(2013)Djolonga, Krause, and Cevher]{NIPS2013_8d34201a}
Djolonga, J., Krause, A., and Cevher, V.
\newblock High-dimensional gaussian process bandits.
\newblock In Burges, C., Bottou, L., Welling, M., Ghahramani, Z., and Weinberger, K. (eds.), \emph{Advances in Neural Information Processing Systems}, volume~26. Curran Associates, Inc., 2013.
\newblock URL \url{https://proceedings.neurips.cc/paper_files/paper/2013/file/8d34201a5b85900908db6cae92723617-Paper.pdf}.

\bibitem[Duembgen(2010)]{Duembgen2010BoundingSG}
Duembgen, L.
\newblock Bounding standard gaussian tail probabilities.
\newblock \emph{arXiv: Statistics Theory}, 2010.
\newblock URL \url{https://api.semanticscholar.org/CorpusID:88512442}.

\bibitem[Duvenaud et~al.(2011)Duvenaud, Nickisch, and Rasmussen]{duvenaud2011additive}
Duvenaud, D.~K., Nickisch, H., and Rasmussen, C.
\newblock Additive gaussian processes.
\newblock In Shawe-Taylor, J., Zemel, R., Bartlett, P., Pereira, F., and Weinberger, K. (eds.), \emph{Advances in Neural Information Processing Systems}, volume~24. Curran Associates, Inc., 2011.
\newblock URL \url{https://proceedings.neurips.cc/paper_files/paper/2011/file/4c5bde74a8f110656874902f07378009-Paper.pdf}.

\bibitem[Eduardo \& Gutmann(2022)Eduardo and Gutmann]{Eduardo2022BayesianOW}
Eduardo, A. and Gutmann, M.~U.
\newblock Bayesian optimization with informative covariance.
\newblock \emph{Trans. Mach. Learn. Res.}, 2023, 2022.
\newblock URL \url{https://api.semanticscholar.org/CorpusID:251320397}.

\bibitem[Eriksson \& Jankowiak(2021)Eriksson and Jankowiak]{pmlr-v161-eriksson21a}
Eriksson, D. and Jankowiak, M.
\newblock High-dimensional {Bayesian} optimization with sparse axis-aligned subspaces.
\newblock In de~Campos, C. and Maathuis, M.~H. (eds.), \emph{Proceedings of the Thirty-Seventh Conference on Uncertainty in Artificial Intelligence}, volume 161 of \emph{Proceedings of Machine Learning Research}, pp.\  493--503. PMLR, 27--30 Jul 2021.
\newblock URL \url{https://proceedings.mlr.press/v161/eriksson21a.html}.

\bibitem[Eriksson et~al.(2019)Eriksson, Pearce, Gardner, Turner, and Poloczek]{eriksson2019turbo}
Eriksson, D., Pearce, M., Gardner, J., Turner, R.~D., and Poloczek, M.
\newblock Scalable global optimization via local bayesian optimization.
\newblock In Wallach, H., Larochelle, H., Beygelzimer, A., d\textquotesingle Alch\'{e}-Buc, F., Fox, E., and Garnett, R. (eds.), \emph{Advances in Neural Information Processing Systems}, volume~32. Curran Associates, Inc., 2019.
\newblock URL \url{https://proceedings.neurips.cc/paper/2019/file/6c990b7aca7bc7058f5e98ea909e924b-Paper.pdf}.

\bibitem[Gardner et~al.(2017)Gardner, Guo, Weinberger, Garnett, and Grosse]{gardner-aistats17a}
Gardner, J., Guo, C., Weinberger, K., Garnett, R., and Grosse, R.
\newblock Discovering and {Exploiting} {Additive} {Structure} for {Bayesian} {Optimization}.
\newblock In Singh, A. and Zhu, J. (eds.), \emph{Proceedings of the Seventeenth International Conference on Artificial Intelligence and Statistics ({AISTATS})}, volume~54, pp.\  1311--1319. Proceedings of Machine Learning Research, 2017.

\bibitem[Garnett et~al.(2014)Garnett, Osborne, and Hennig]{garnett2014active}
Garnett, R., Osborne, M.~A., and Hennig, P.
\newblock Active learning of linear embeddings for gaussian processes.
\newblock In \emph{Proceedings of the Thirtieth Conference on Uncertainty in Artificial Intelligence}, UAI'14, pp.\  230–239, Arlington, Virginia, USA, 2014. AUAI Press.
\newblock ISBN 9780974903910.

\bibitem[GPyOpt-authors(2016)]{gpyopt2016}
GPyOpt-authors, T.
\newblock {GPyOpt}: A bayesian optimization framework in python.
\newblock \url{http://github.com/SheffieldML/GPyOpt}, 2016.

\bibitem[Griffiths \& Hern{\'a}ndez-Lobato(2017)Griffiths and Hern{\'a}ndez-Lobato]{Griffiths2017ConstrainedBO}
Griffiths, R.-R. and Hern{\'a}ndez-Lobato, J.~M.
\newblock Constrained bayesian optimization for automatic chemical design.
\newblock \emph{arXiv: Machine Learning}, 2017.

\bibitem[Han et~al.(2021)Han, Arora, and Scarlett]{Han_Arora_Scarlett_2021}
Han, E., Arora, I., and Scarlett, J.
\newblock High-dimensional bayesian optimization via tree-structured additive models.
\newblock \emph{Proceedings of the AAAI Conference on Artificial Intelligence}, 35\penalty0 (9):\penalty0 7630--7638, May 2021.
\newblock URL \url{https://ojs.aaai.org/index.php/AAAI/article/view/16933}.

\bibitem[Hansen(2006)]{hansen-eda06a}
Hansen, N.
\newblock The {CMA} evolution strategy: a comparing review.
\newblock In Lozano, J., Larranaga, P., Inza, I., and Bengoetxea, E. (eds.), \emph{Towards a new evolutionary computation. Advances on estimation of distribution algorithms}, pp.\  75--102. 2006.

\bibitem[Hellsten et~al.(2023)Hellsten, Hvarfner, Papenmeier, and Nardi]{hellsten2023highdimensional}
Hellsten, E.~O., Hvarfner, C., Papenmeier, L., and Nardi, L.
\newblock High-dimensional bayesian optimization with group testing, 2023.

\bibitem[Hutter et~al.(2011)Hutter, Hoos, and Leyton-Brown]{smac}
Hutter, F., Hoos, H.~H., and Leyton-Brown, K.
\newblock Sequential model-based optimization for general algorithm configuration.
\newblock In \emph{Learning and Intelligent Optimization}, 2011.

\bibitem[Hvarfner et~al.(2023)Hvarfner, Hellsten, Hutter, and Nardi]{hvarfner2023selfcorrecting}
Hvarfner, C., Hellsten, E.~O., Hutter, F., and Nardi, L.
\newblock Self-correcting bayesian optimization through bayesian active learning.
\newblock In \emph{Thirty-seventh Conference on Neural Information Processing Systems}, 2023.
\newblock URL \url{https://openreview.net/forum?id=dX9MjUtP1A}.

\bibitem[Jones et~al.(1998)Jones, Schonlau, and Welch]{jones-jgo98a}
Jones, D., Schonlau, M., and Welch, W.
\newblock Efficient global optimization of expensive black box functions.
\newblock 13:\penalty0 455--492, 1998.

\bibitem[Jones(2001)]{jones-jgo01a}
Jones, D.~R.
\newblock A taxonomy of global optimization methods based on response surfaces.
\newblock 21:\penalty0 345--383, 2001.

\bibitem[Kandasamy et~al.(2015)Kandasamy, Schneider, and Póczos]{kandasamy-icml15a}
Kandasamy, K., Schneider, J., and Póczos, B.
\newblock High {Dimensional} {Bayesian} {Optimisation} and {Bandits} via {Additive} {Models}.
\newblock In Bach, F. and Blei, D. (eds.), \emph{Proceedings of the 32nd International Conference on Machine Learning ({ICML}'15)}, volume~37, pp.\  295--304. Omnipress, 2015.

\bibitem[Karvonen \& Oates(2023)Karvonen and Oates]{JMLR:v24:22-1153}
Karvonen, T. and Oates, C.~J.
\newblock Maximum likelihood estimation in gaussian process regression is ill-posed.
\newblock \emph{Journal of Machine Learning Research}, 24\penalty0 (120):\penalty0 1--47, 2023.
\newblock URL \url{http://jmlr.org/papers/v24/22-1153.html}.

\bibitem[Kirschner et~al.(2019)Kirschner, Mutny, Hiller, Ischebeck, and Krause]{pmlr-v97-kirschner19a}
Kirschner, J., Mutny, M., Hiller, N., Ischebeck, R., and Krause, A.
\newblock Adaptive and safe {B}ayesian optimization in high dimensions via one-dimensional subspaces.
\newblock In Chaudhuri, K. and Salakhutdinov, R. (eds.), \emph{Proceedings of the 36th International Conference on Machine Learning}, volume~97 of \emph{Proceedings of Machine Learning Research}, pp.\  3429--3438. PMLR, 09--15 Jun 2019.
\newblock URL \url{https://proceedings.mlr.press/v97/kirschner19a.html}.

\bibitem[K{\"o}ppen(2000)]{koppencurse}
K{\"o}ppen, M.
\newblock The curse of dimensionality.
\newblock \emph{5th online world conference on soft computing in industrial applications (WSC5)}, 2000.

\bibitem[Krause et~al.(2008)Krause, Singh, and Guestrin]{JMLR:v9:krause08a}
Krause, A., Singh, A., and Guestrin, C.
\newblock Near-optimal sensor placements in gaussian processes: Theory, efficient algorithms and empirical studies.
\newblock \emph{Journal of Machine Learning Research}, 9\penalty0 (8):\penalty0 235--284, 2008.
\newblock URL \url{http://jmlr.org/papers/v9/krause08a.html}.

\bibitem[Letham et~al.(2020)Letham, Calandra, Rai, and Bakshy]{letham2020re}
Letham, B., Calandra, R., Rai, A., and Bakshy, E.
\newblock {Re-examining linear embeddings for high-dimensional Bayesian optimization}.
\newblock \emph{Advances in neural information processing systems}, 33:\penalty0 1546--1558, 2020.

\bibitem[Li et~al.(2017)Li, Gupta, Rana, Nguyen, Venkatesh, and Shilton]{li2017dropout}
Li, C., Gupta, S., Rana, S., Nguyen, V., Venkatesh, S., and Shilton, A.
\newblock High dimensional bayesian optimization using dropout.
\newblock In \emph{Proceedings of the 26th International Joint Conference on Artificial Intelligence}, IJCAI'17, pp.\  2096–2102. AAAI Press, 2017.
\newblock ISBN 9780999241103.

\bibitem[Malu et~al.(2021)Malu, Dasarathy, and Spanias]{maju2021hdsurvey}
Malu, M., Dasarathy, G., and Spanias, A.
\newblock Bayesian optimization in high-dimensional spaces: A brief survey.
\newblock In \emph{2021 12th International Conference on Information, Intelligence, Systems and Applications (IISA)}, pp.\  1--8, 2021.
\newblock \doi{10.1109/IISA52424.2021.9555522}.

\bibitem[Maus et~al.(2022)Maus, Jones, Moore, Kusner, Bradshaw, and Gardner]{NEURIPS2022_ded98d28}
Maus, N., Jones, H., Moore, J., Kusner, M.~J., Bradshaw, J., and Gardner, J.
\newblock Local latent space bayesian optimization over structured inputs.
\newblock In Koyejo, S., Mohamed, S., Agarwal, A., Belgrave, D., Cho, K., and Oh, A. (eds.), \emph{Advances in Neural Information Processing Systems}, volume~35, pp.\  34505--34518. Curran Associates, Inc., 2022.

\bibitem[Moore et~al.(2016)Moore, Chua, Berry, and Gair]{Moore_2016}
Moore, C.~J., Chua, A. J.~K., Berry, C. P.~L., and Gair, J.~R.
\newblock Fast methods for training gaussian processes on large datasets.
\newblock \emph{Royal Society Open Science}, 3\penalty0 (5):\penalty0 160125, May 2016.
\newblock ISSN 2054-5703.
\newblock \doi{10.1098/rsos.160125}.
\newblock URL \url{http://dx.doi.org/10.1098/rsos.160125}.

\bibitem[M{\"u}ller et~al.(2021)M{\"u}ller, von Rohr, and Trimpe]{GIBO}
M{\"u}ller, S., von Rohr, A., and Trimpe, S.
\newblock Local policy search with bayesian optimization.
\newblock In \emph{Advances in Neural Information Processing Systems}, 2021.

\bibitem[Nayebi et~al.(2019)Nayebi, Munteanu, and Poloczek]{nayebi2019framework}
Nayebi, A., Munteanu, A., and Poloczek, M.
\newblock {A framework for Bayesian optimization in embedded subspaces}.
\newblock In \emph{International Conference on Machine Learning}, pp.\  4752--4761. PMLR, 2019.

\bibitem[Nemhauser et~al.(1978)Nemhauser, Wolsey, and Fisher]{nemhauser1978submod}
Nemhauser, G.~L., Wolsey, L.~A., and Fisher, M.~L.
\newblock An analysis of approximations for maximizing submodular set functions--i.
\newblock \emph{Math. Program.}, 14\penalty0 (1):\penalty0 265–294, dec 1978.
\newblock ISSN 0025-5610.
\newblock \doi{10.1007/BF01588971}.
\newblock URL \url{https://doi.org/10.1007/BF01588971}.

\bibitem[Nguyen et~al.(2022)Nguyen, Wu, Gardner, and Garnett]{NEURIPS2022_555479a2}
Nguyen, Q., Wu, K., Gardner, J., and Garnett, R.
\newblock Local bayesian optimization via maximizing probability of descent.
\newblock In Koyejo, S., Mohamed, S., Agarwal, A., Belgrave, D., Cho, K., and Oh, A. (eds.), \emph{Advances in Neural Information Processing Systems}, volume~35, pp.\  13190--13202. Curran Associates, Inc., 2022.

\bibitem[Ober et~al.(2021)Ober, Rasmussen, and van~der Wilk]{pmlr-v161-ober21a}
Ober, S.~W., Rasmussen, C.~E., and van~der Wilk, M.
\newblock The promises and pitfalls of deep kernel learning.
\newblock In de~Campos, C. and Maathuis, M.~H. (eds.), \emph{Proceedings of the Thirty-Seventh Conference on Uncertainty in Artificial Intelligence}, volume 161 of \emph{Proceedings of Machine Learning Research}, pp.\  1206--1216. PMLR, 27--30 Jul 2021.
\newblock URL \url{https://proceedings.mlr.press/v161/ober21a.html}.

\bibitem[Oh et~al.(2018)Oh, Gavves, and Welling]{oh18bock}
Oh, C., Gavves, E., and Welling, M.
\newblock {BOCK} : {B}ayesian optimization with cylindrical kernels.
\newblock In Dy, J. and Krause, A. (eds.), \emph{Proceedings of the 35th International Conference on Machine Learning}, volume~80 of \emph{Proceedings of Machine Learning Research}, pp.\  3868--3877. PMLR, 10--15 Jul 2018.
\newblock URL \url{https://proceedings.mlr.press/v80/oh18a.html}.

\bibitem[Osborne(2010)]{osborne2010bayesian}
Osborne, M.~A.
\newblock \emph{Bayesian Gaussian processes for sequential prediction, optimisation and quadrature}.
\newblock PhD thesis, Oxford University, UK, 2010.

\bibitem[Papenmeier et~al.(2022)Papenmeier, Nardi, and Poloczek]{papenmeier2022increasing}
Papenmeier, L., Nardi, L., and Poloczek, M.
\newblock Increasing the scope as you learn: Adaptive bayesian optimization in nested subspaces.
\newblock In Oh, A.~H., Agarwal, A., Belgrave, D., and Cho, K. (eds.), \emph{Advances in Neural Information Processing Systems}, 2022.
\newblock URL \url{https://openreview.net/forum?id=e4Wf6112DI}.

\bibitem[Papenmeier et~al.(2023)Papenmeier, Nardi, and Poloczek]{papenmeier2023bounce}
Papenmeier, L., Nardi, L., and Poloczek, M.
\newblock {Bounce: a Reliable Bayesian Optimization Algorithm for Combinatorial and Mixed Spaces}.
\newblock \emph{arXiv preprint arXiv:2307.00618}, 2023.

\bibitem[Rana et~al.(2017)Rana, Li, Gupta, Nguyen, and Venkatesh]{rana2017elastic}
Rana, S., Li, C., Gupta, S., Nguyen, V., and Venkatesh, S.
\newblock High dimensional {B}ayesian optimization with elastic {G}aussian process.
\newblock In Precup, D. and Teh, Y.~W. (eds.), \emph{Proceedings of the 34th International Conference on Machine Learning}, volume~70 of \emph{Proceedings of Machine Learning Research}, pp.\  2883--2891. PMLR, 06--11 Aug 2017.
\newblock URL \url{https://proceedings.mlr.press/v70/rana17a.html}.

\bibitem[Rasmussen \& Williams(2006)Rasmussen and Williams]{rasmussen-book06a}
Rasmussen, C. and Williams, C.
\newblock \emph{Gaussian Processes for Machine Learning}.
\newblock The MIT Press, 2006.

\bibitem[Rolland et~al.(2018)Rolland, Scarlett, Bogunovic, and Cevher]{pmlr-v84-rolland18a}
Rolland, P., Scarlett, J., Bogunovic, I., and Cevher, V.
\newblock High-dimensional bayesian optimization via additive models with overlapping groups.
\newblock In Storkey, A. and Perez-Cruz, F. (eds.), \emph{Proceedings of the Twenty-First International Conference on Artificial Intelligence and Statistics}, volume~84 of \emph{Proceedings of Machine Learning Research}, pp.\  298--307. PMLR, 09--11 Apr 2018.
\newblock URL \url{https://proceedings.mlr.press/v84/rolland18a.html}.

\bibitem[Rothfuss et~al.(2021{\natexlab{a}})Rothfuss, Fortuin, Josifoski, and Krause]{pmlr-v139-rothfuss21a}
Rothfuss, J., Fortuin, V., Josifoski, M., and Krause, A.
\newblock Pacoh: Bayes-optimal meta-learning with pac-guarantees.
\newblock In Meila, M. and Zhang, T. (eds.), \emph{Proceedings of the 38th International Conference on Machine Learning}, volume 139 of \emph{Proceedings of Machine Learning Research}, pp.\  9116--9126. PMLR, 18--24 Jul 2021{\natexlab{a}}.
\newblock URL \url{https://proceedings.mlr.press/v139/rothfuss21a.html}.

\bibitem[Rothfuss et~al.(2021{\natexlab{b}})Rothfuss, Heyn, Chen, and Krause]{rothfuss2021fpacoh}
Rothfuss, J., Heyn, D., Chen, J., and Krause, A.
\newblock Meta-learning reliable priors in the function space.
\newblock In \emph{Advances in Neural Information Processing Systems}, volume~34, 2021{\natexlab{b}}.

\bibitem[{\v{S}}ehi{\'c} et~al.(2021){\v{S}}ehi{\'c}, Gramfort, Salmon, and Nardi]{vsehic2021lassobench}
{\v{S}}ehi{\'c}, K., Gramfort, A., Salmon, J., and Nardi, L.
\newblock Lasso{B}ench: {A} {H}igh-{D}imensional {H}yperparameter {O}ptimization {B}enchmark {S}uite for {L}asso.
\newblock \emph{arXiv preprint arXiv:2111.02790}, 2021.

\bibitem[Siivola et~al.(2017)Siivola, Vehtari, Vanhatalo, Gonz{\'a}lez, and Andersen]{Siivola2017CORRECTINGBO}
Siivola, E., Vehtari, A., Vanhatalo, J.~P., Gonz{\'a}lez, J.~I., and Andersen, M.~R.
\newblock Correcting boundary over-exploration deficiencies in bayesian optimization with virtual derivative sign observations.
\newblock \emph{2018 IEEE 28th International Workshop on Machine Learning for Signal Processing (MLSP)}, pp.\  1--6, 2017.
\newblock URL \url{https://api.semanticscholar.org/CorpusID:53236252}.

\bibitem[Snoek et~al.(2012{\natexlab{a}})Snoek, Larochelle, and Adams]{snoek-nips12a}
Snoek, J., Larochelle, H., and Adams, R.
\newblock Practical {B}ayesian optimization of machine learning algorithms.
\newblock In Bartlett, P., Pereira, F., Burges, C., Bottou, L., and Weinberger, K. (eds.), \emph{Proceedings of the 26th International Conference on Advances in Neural Information Processing Systems ({N}eur{IPS}'12)}, pp.\  2960--2968, 2012{\natexlab{a}}.

\bibitem[Snoek et~al.(2012{\natexlab{b}})Snoek, Larochelle, and Adams]{spearmint}
Snoek, J., Larochelle, H., and Adams, R.~P.
\newblock Practical bayesian optimization of machine learning algorithms.
\newblock Advances in Neural Information Processing Systems, pp.\  2951–2959, Red Hook, NY, USA, 2012{\natexlab{b}}. Curran Associates Inc.

\bibitem[Snoek et~al.(2014)Snoek, Swersky, Zemel, and Adams]{snoek-icml14a}
Snoek, J., Swersky, K., Zemel, R., and Adams, R.
\newblock Input warping for {Bayesian} optimization of non-stationary functions.
\newblock In Xing, E. and Jebara, T. (eds.), \emph{Proceedings of the 31th International Conference on Machine Learning, ({ICML}'14)}, pp.\  1674--1682. Omnipress, 2014.

\bibitem[Song et~al.(2022)Song, Xue, Huang, and Qian]{song2022monte}
Song, L., Xue, K., Huang, X., and Qian, C.
\newblock Monte carlo tree search based variable selection for high dimensional bayesian optimization.
\newblock In Oh, A.~H., Agarwal, A., Belgrave, D., and Cho, K. (eds.), \emph{Advances in Neural Information Processing Systems}, 2022.
\newblock URL \url{https://openreview.net/forum?id=SUzPos_pUC}.

\bibitem[Srinivas et~al.(2010)Srinivas, Krause, Kakade, and Seeger]{srninivas-icml10a}
Srinivas, N., Krause, A., Kakade, S., and Seeger, M.
\newblock {G}aussian process optimization in the bandit setting: No regret and experimental design.
\newblock In F{\"u}rnkranz, J. and Joachims, T. (eds.), \emph{Proceedings of the 27th International Conference on Machine Learning ({ICML}'10)}, pp.\  1015--1022. Omnipress, 2010.

\bibitem[Srinivas et~al.(2012)Srinivas, Krause, Kakade, and Seeger]{Srinivas_2012}
Srinivas, N., Krause, A., Kakade, S.~M., and Seeger, M.~W.
\newblock Information-theoretic regret bounds for gaussian process optimization in the bandit setting.
\newblock \emph{IEEE Transactions on Information Theory}, 58\penalty0 (5):\penalty0 3250–3265, May 2012.
\newblock ISSN 1557-9654.
\newblock \doi{10.1109/tit.2011.2182033}.
\newblock URL \url{http://dx.doi.org/10.1109/TIT.2011.2182033}.

\bibitem[Swersky et~al.(2014)Swersky, Duvenaud, Snoek, Hutter, and Osborne]{swersky2014raiders}
Swersky, K., Duvenaud, D., Snoek, J., Hutter, F., and Osborne, M.~A.
\newblock Raiders of the lost architecture: Kernels for bayesian optimization in conditional parameter spaces, 2014.

\bibitem[Swersky(2017)]{swersky2017improving}
Swersky, K.~J.
\newblock \emph{Improving Bayesian Optimization for Machine Learning using Expert Priors}.
\newblock PhD thesis, 2017.

\bibitem[Todorov et~al.(2012)Todorov, Erez, and Tassa]{todorov2012mujoco}
Todorov, E., Erez, T., and Tassa, Y.
\newblock Mujoco: A physics engine for model-based control.
\newblock In \emph{2012 IEEE/RSJ International Conference on Intelligent Robots and Systems}, pp.\  5026--5033. IEEE, 2012.
\newblock \doi{10.1109/IROS.2012.6386109}.

\bibitem[Wan et~al.(2021)Wan, Nguyen, Ha, Ru, Lu, and Osborne]{pmlr-v139-wan21b}
Wan, X., Nguyen, V., Ha, H., Ru, B., Lu, C., and Osborne, M.~A.
\newblock Think global and act local: Bayesian optimisation over high-dimensional categorical and mixed search spaces.
\newblock In Meila, M. and Zhang, T. (eds.), \emph{Proceedings of the 38th International Conference on Machine Learning}, volume 139 of \emph{Proceedings of Machine Learning Research}, pp.\  10663--10674. PMLR, 18--24 Jul 2021.
\newblock URL \url{https://proceedings.mlr.press/v139/wan21b.html}.

\bibitem[Wang et~al.(2020)Wang, Fonseca, and Tian]{NEURIPS2020_e2ce14e8}
Wang, L., Fonseca, R., and Tian, Y.
\newblock Learning search space partition for black-box optimization using monte carlo tree search.
\newblock In Larochelle, H., Ranzato, M., Hadsell, R., Balcan, M., and Lin, H. (eds.), \emph{Advances in Neural Information Processing Systems}, volume~33, pp.\  19511--19522. Curran Associates, Inc., 2020.
\newblock URL \url{https://proceedings.neurips.cc/paper_files/paper/2020/file/e2ce14e81dba66dbff9cbc35ecfdb704-Paper.pdf}.

\bibitem[Wang et~al.(2016)Wang, Hutter, Zoghi, Matheson, and De~Feitas]{wang2016bayesian}
Wang, Z., Hutter, F., Zoghi, M., Matheson, D., and De~Feitas, N.
\newblock Bayesian optimization in a billion dimensions via random embeddings.
\newblock \emph{Journal of Artificial Intelligence Research}, 55:\penalty0 361--387, 2016.

\bibitem[Wang et~al.(2017)Wang, Li, Jegelka, and Kohli]{pmlr-v70-wang17h}
Wang, Z., Li, C., Jegelka, S., and Kohli, P.
\newblock Batched high-dimensional {B}ayesian optimization via structural kernel learning.
\newblock In Precup, D. and Teh, Y.~W. (eds.), \emph{Proceedings of the 34th International Conference on Machine Learning}, volume~70 of \emph{Proceedings of Machine Learning Research}, pp.\  3656--3664. PMLR, 06--11 Aug 2017.
\newblock URL \url{https://proceedings.mlr.press/v70/wang17h.html}.

\bibitem[Williams \& Rasmussen(1995)Williams and Rasmussen]{NIPS1995_7cce53cf}
Williams, C. and Rasmussen, C.
\newblock Gaussian processes for regression.
\newblock In Touretzky, D., Mozer, M., and Hasselmo, M. (eds.), \emph{Advances in Neural Information Processing Systems}, volume~8. MIT Press, 1995.
\newblock URL \url{https://proceedings.neurips.cc/paper_files/paper/1995/file/7cce53cf90577442771720a370c3c723-Paper.pdf}.

\bibitem[Wu et~al.(2023)Wu, Kim, Garnett, and Gardner]{wu2023the}
Wu, K., Kim, K., Garnett, R., and Gardner, J.~R.
\newblock The behavior and convergence of local bayesian optimization.
\newblock In \emph{Thirty-seventh Conference on Neural Information Processing Systems}, 2023.
\newblock URL \url{https://openreview.net/forum?id=9KtX12YmA7}.

\bibitem[Yao et~al.(2020)Yao, Vehtari, and Gelman]{yao2020stacking}
Yao, Y., Vehtari, A., and Gelman, A.
\newblock Stacking for non-mixing bayesian computations: The curse and blessing of multimodal posteriors, 2020.

\bibitem[Ziomek \& Bou~Ammar(2023)Ziomek and Bou~Ammar]{pmlr-v202-ziomek23a}
Ziomek, J.~K. and Bou~Ammar, H.
\newblock Are random decompositions all we need in high dimensional {B}ayesian optimisation?
\newblock In Krause, A., Brunskill, E., Cho, K., Engelhardt, B., Sabato, S., and Scarlett, J. (eds.), \emph{Proceedings of the 40th International Conference on Machine Learning}, volume 202 of \emph{Proceedings of Machine Learning Research}, pp.\  43347--43368. PMLR, 23--29 Jul 2023.
\newblock URL \url{https://proceedings.mlr.press/v202/ziomek23a.html}.

\end{thebibliography}
\cleardoublepage

\newpage
\appendix
\onecolumn
\section{Experimental Setup}\label{app:expsetup}
We outline our experimental setup, detailing the specifics of the baseline methods. Our code is publically available at~\url{https://github.com/hvarfner/vanilla_bo_in_highdim}.

\subsection{SAASBO}\label{sec:saasbosetup}
Due to the computational expense of SAASBO model fitting and subsequent forward passes through the acquisition function, we run the lower-budget variant outlined in App. A1 of~\citep{pmlr-v161-eriksson21a}. Due to the 16-fold increase in the cost of the forward pass in the fully Bayesian setting, we limit the maximal number of evaluations to 500. Lastly, the model is retrained once every 4 iterations, since a single model fitting takes upwards of 5 minutes on SVM on 4 CPUS for later iterations.

\subsection{Complexity Comparison Setup}\label{app:migcomp}
For the complexity comparison in Fig.~\ref{fig:hdcomp}, we benchmark against the following instantiations of each HDBO algorithm:

\paragraph{Cylindrical Kernel} We stay as close as possible to the default values presented in BOCK~\cite{oh18bock}, using a degree-3 polynomial for the angular component of the kernel. We disregard input warpings~\cite{snoek-icml14a}, as we do not view them as related to the dimensionality-reducing component of the model.

\paragraph{AddGP} We randomly sample groups by sequentially considering one dimension at a time. With equal probability for each existing group and one empty group, the dimension under consideration is added to one of the groups. This setup is similar, but not equivalent, to the algorithm by~\citet{pmlr-v202-ziomek23a}.

\paragraph{Local GP} We consider the instantiation described by~\citet{eriksson2019turbo} of constructing the trust region with a size of $0.8\bm{\ell}$. Since we plot the complexity of the local GP after one round of shrinkage, the trust region side length is equal to $0.4\bm{\ell}$. Moreover, we assume this trust region to hold \textit{all} data to enable a comparison with other methods.

\paragraph{Random Embedding} We instantiate the standard REMBO~\cite{wang2016bayesian} embedding from a $D$-dimensional Gaussian distribution. We compute the complexity by based on the shortest lengthscale of each subspace dimension. These, in turn, are computed from by multiplying the base lengthscale, $\ell = 0.5$, by the fraction of the effective dimension that the ambient dimension is occupying, as visualized in Fig.3 of~\citet{wang2016bayesian}.

\section{Benchmarks}\label{app:benches}
We outline the exact benchmarks used, including the exact bounds and the noise level. For Levy, we offset each active dimension. Since the optimum is in $[1,1,1,\ldots]$, methods that search along diagonals in the search space, such as BAxUS~\cite{papenmeier2022increasing}, gets an outsized, and arguably unrealistic, advantage. 

Notably, We use the BAxUS variant of SVM, which uses a smaller subset than the original benchmark by~\cite{pmlr-v161-eriksson21a} for faster evaluation. As evident form a comparison of the results in~\citet{papenmeier2022increasing} and ~\cite{pmlr-v161-eriksson21a}, these benchmarks produce different results, suggesting that the objective functions are substantially different.

\begin{table}[h!]
    \centering
    \begin{tabular}{c|c|c|c}
    \hline
    Task & Effective Dimensionality & $\sigma_\epsilon$& Search space  \\
    \hline
    Levy (4D) & $4$ & $0.01$ & $[-10, 5]\times[-10, 10]\times[-5, 10]\times[-1, 10]$\\
    Hartmann (6D) & $6$ & $0.01$ & $[0, 1]^D$\\
    \hline
    \end{tabular}
    \caption{Benchmarks used for the synthetic, axis-aligned experiments.}
    \label{tab:syn_bench}
\end{table}

\begin{table}[h!]
    \centering
    \begin{tabular}{c|c|c}
    \hline
    Task & Dimensionality & Search space  \\
    \hline
    MOPTA & $124$ & $[0, 1]^D$\\
    Lasso-DNA & $180$ & $[-1, 1]^D$\\
    SVM & $388$ & $[0, 1]^D$\\
    Ant & $888$ & $[-1, 1]^D$\\
    Humanoid & $6392$ & $[-1, 1]^D$\\
    \hline
    \end{tabular}
    \caption{Benchmarks used for the real-world optimization experiments.}
    \label{tab:bo_bench}
\end{table}

\section{The Locality Issue}\label{app:boundary}
We outline our findings related to the boundary and locality issue, proving that the original description~\cite{swersky2017improving} does not happen in practice. Subsequently, we display our empirical findings regarding the locality issue for BO with a conventional, short-lengthscale hyperprior using the BoTorch~\cite{balandat2020botorch} default. We note, though, that other BO frameworks~\cite{smac, spearmint, gpyopt2016} employ lengthscale priors with very similar characteristics, and any of them could have been used for our purpose.

\subsection{Theory: Non-existence of the Boundary Issue}\label{app:proof_boundary}

To prove that \ei{} will never prefer a point with maximal variance under the proposed setup, we first restate the formal definition of \ei{}: 

\begin{equation}
    \ei{}(\bm{x}) = \mathbb{E}\left[[(f(\bm{x}) - y_\text{max}]^+\right] = Z\sigma(\bm{x})\Phi(Z) + \sigma(\bm{x})\phi(Z),
\end{equation}
where $y_{max}$ is the maximal objective function value observed so far, $Z = (\mu(\bm{x}) - y_\text{max})/\sigma(\bm{x})$, and $\phi$ and $\Phi$ are the PDF and CDF of the standard Gaussian, respectively. Hence, for a noiseless function the next point $\bm{x}_*$ to query is
\begin{equation}
    \bm{x}_* \in \argmax_{\bm{x}\in\mathcal{X}} \ei{}(\bm{x}).
\end{equation}\label{eq:ei}
Note that we are considering a maximization problem. Lastly, we denote the mean of the GP by $c$.

    

\begin{proposition} 
Assume that $y_\text{max} > c$, $\mathbf{K} = \bm{I}$ and that the candidate query $\bm{x}_*$ correlates with at most one observation. Then, the correlation $\rho^* = \sigma_f^{-2}k(\bm{x}_*, \bm{x}_\text{inc})$ between the next query $\bm{x}_* =\argmax_{\bm{x}\in \mathcal{X}} \ei{}(\bm{x})$ and $\bm{x}_\text{inc}$ satisfies 
\begin{equation}\label{eq:boundproof}
    \rho^*\sqrt{\frac{1+\rho^*}{1-\rho^*}} \geq \frac{y_\text{max} - c}{\sigma_f}.
\end{equation}
\end{proposition}

\emph{Proof}. Since $\mathbf{K} = \sigma_f^2\bm{I}$ and $\bm{x}_*$ correlates with only one observation, we need only consider the proximity of $\bm{x}_*$ to the incumbent $\bm{x}_{inc}$. 
As such, we start by parametrizing \ei{} by the correlation $\rho$ , so that $k(\bm{x}_*, \bm{x}_{inc}) = \sigma_f^2\rho$:

\begin{equation*}
    \mu(\bm{x}_*) = c + k(\bm{x}_*, \bm{X}) k(\bm{X}, \bm{X})^{-1} (\bm{y} - c) = c + \rho(y_\text{max} - c)
\end{equation*} 

\begin{equation}
    \sigma^2(\bm{x}_*) = k(\bm{x}_*, \bm{x}_*) + k(\bm{x}_*, \bm{X}) k(\bm{X}, \bm{X})^{-1} k(\bm{X}, \bm{x}_*) = \sigma_f^2(1-\rho^2).
\end{equation} 
We will by $\hat{\ei{}}(\rho)$ denote the $\rho$-parametrized $\ei{}$, which is a function over $\rho$, to distinguish it from the regular $\ei{}(\bm{x})$, which is a function over $\bm{x}$. Moreover, for ease of notation, we scale our data such that $c = 0$, $\sigma_f^2 = 1$, and $\hat{y}_\text{max} = \frac{y_\text{max}-c}{\sigma_f} > 0$. After scaling we further get $\hat{Z} = \left(\frac{\left(\rho - 1\right)}{\sqrt{1 - \rho^{2}}}\hat{y}_\text{max}\right)
$, and finally:

\begin{equation}
    \hat{\ei{}}(\rho) =\left(\rho - 1\right) \hat{y}_\text{max} \Phi\left(\frac{\left(\rho - 1\right)}{\sqrt{1 - \rho^{2}}}\hat{y}_\text{max}\right) + \sqrt{1 - \rho^{2}}\phi\left(\frac{\left(\rho - 1\right)}{\sqrt{1 - \rho^{2}}}\hat{y}_\text{max}\right).
\end{equation}


Differentiating with regard to $\rho$, we obtain that:
\begin{equation}
    \frac{\partial\hat{\ei{}}}{\partial\rho} =  \hat{y}_\text{max}\Phi(\hat{Z}) - \frac{\rho}{\sqrt{1-\rho^2}}\phi(\hat{Z}).
\end{equation}



We then establish the values of $\rho$ for which $\frac{\partial\ei{}}{\partial\rho}$ are positive. We bound $\Phi(\hat{Z})$ as a function of $\phi(\hat{Z})$ using known inequalities for Mills' ratio $(1-\Phi)/\phi$~\cite{Duembgen2010BoundingSG},
\begin{equation*}
    \frac{2\phi(\hat{Z})}{\sqrt{4 + \hat{Z}^2}-\hat{Z}}\leq \Phi(\hat{Z}), \quad -\hat{y}_\text{max} \le \hat{Z} \leq 0. 
\end{equation*}
We then insert the inequality into the expression for $\frac{\partial\hat{\ei{}}}{\partial\rho}$, which leaves
\begin{align*}
    \frac{\partial\hat{\ei{}}}{\partial\rho} &\geq \left[\frac{2\hat{y}_\text{max}}{\sqrt{4 + \hat{Z}^2} - \hat{Z}}-\frac{\rho}{\sqrt{1-\rho^2}}\right]\phi(\hat{Z})\\
    &= \left[\frac{2\hat{y}_\text{max}}{\sqrt{4 + \frac{\left(1-\rho\right)^2}{1 - \rho^{2}}\hat{y}_\text{max}^2} + \frac{1-\rho}{\sqrt{1 - \rho^{2}}}\hat{y}_\text{max}}-\frac{\rho}{\sqrt{1-\rho^2}}\right]\phi(\hat{Z})\\
    &\geq \left[\frac{2\hat{y}_\text{max}}{{2 + 2\hat{y}_\text{max}\frac{1-\rho}{\sqrt{1 - \rho^{2}}}}}-\frac{\rho}{\sqrt{1-\rho^2}}\right]\phi(\hat{Z}),
\end{align*}

where we apply the triangle inequality to the denominator of the left term between the first and second step. We wish to lower bound $\rho^* = \argmax \hat{\ei}(\rho)$ by finding the values of $\rho$ for which $\hat{\ei}$ is strictly increasing. Thus, we set our lower bound for the derivative to zero and obtain
\begin{equation*}
    \frac{\partial\hat{\ei{}}}{\partial\rho} \geq \left[\frac{\hat{y}_\text{max}}{{1 + \hat{y}_\text{max}\frac{1-\rho}{\sqrt{1 - \rho^{2}}}}}-\frac{\rho}{\sqrt{1-\rho^2}}\right]\phi(\hat{Z}) \geq 0.
\end{equation*}
We then multiply by $\sqrt{1 - \rho^{2}}$, and divide by $\phi(\hat{Z})$, which are both $>0$ for $\rho > 0$, and re-arrange:
\begin{align*}
    & \frac{\hat{y}_\text{max}(1-p^2)}{{\sqrt{1-\rho^2} + \hat{y}_\text{max}(1-\rho)}}-\rho \geq 0 \iff\\
    & \frac{\hat{y}_\text{max}(1+\rho)}{\sqrt{\frac{1+\rho}{1-\rho}}+\hat{y}_\text{max}}-\rho \geq 0\iff\\
    & {\hat{y}_\text{max}} - \rho{\sqrt{\frac{1+\rho}{1-\rho}}} \geq 0
\end{align*}
As such, we can conclude, that $\hat{\ei{}}$ is strictly increasing in $\rho$ as long as ${\hat{y}_\text{max}} \geq \rho{\sqrt{\frac{1+\rho}{1-\rho}}}$, meaning that $\rho^* = \argmax_{\rho \in [0, 1)} \hat{\ei}({\rho})$ satisfies the inequality
\begin{equation*}
    \rho^*\sqrt{\frac{1+\rho^*}{1-\rho^*}} \geq \hat{y}_\text{max} = \frac{y_\text{max} -c}{\sigma_f}.
\end{equation*}
\qed

\subsection{The Locality Issue in Practice}\label{app:boundary_emp}
We now demonstrate the practical prevalence of the locality issue, expanding on the claims from Sec.~\ref{sec:boundary}. We note that the locality issue applies to the high-complexity setting, as generally outlined in Sec.~\ref{sec:issues}. While the high-complexity setting is most prevalent when modeling high-dimensional problems, we will exemplify the locality issue through low-dimensional examples where complexity has been artificially increased by shortening the lengthscale of the model.

\begin{figure}
    \centering
\begin{minipage}[b]{0.28\textwidth}
    \includegraphics[width=\linewidth]{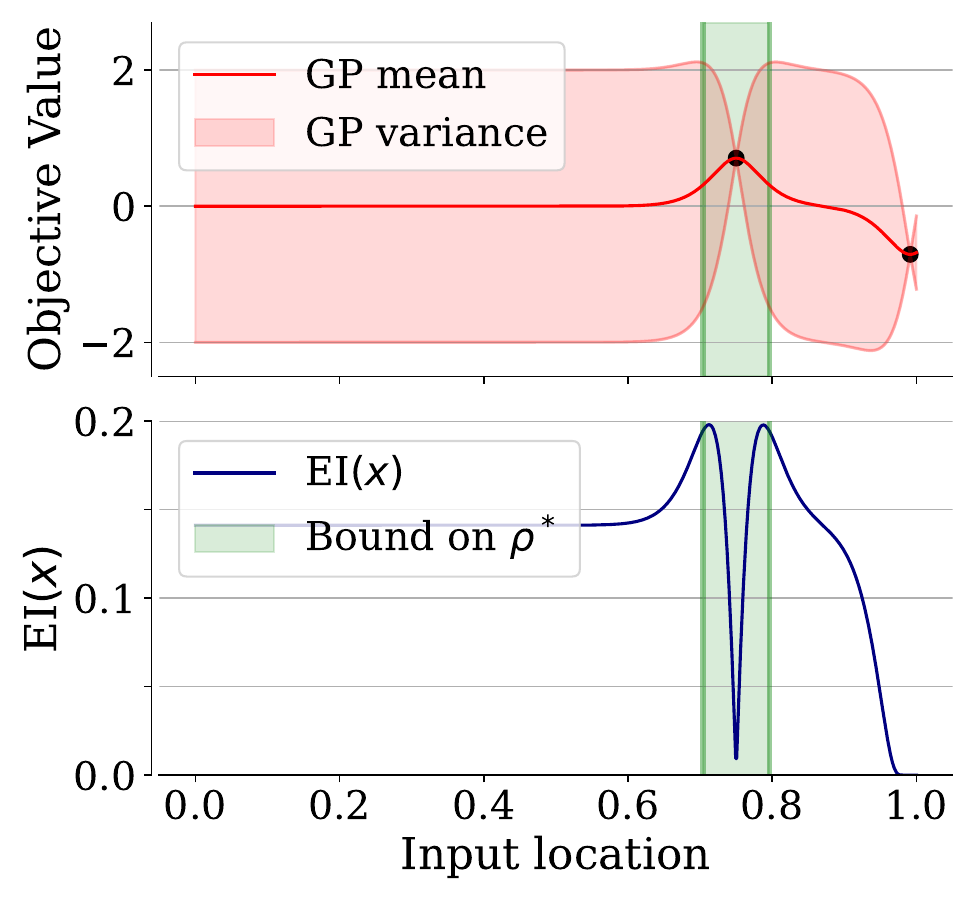}
\end{minipage}
\begin{minipage}[b]{0.28\textwidth}
    \includegraphics[width=\linewidth]{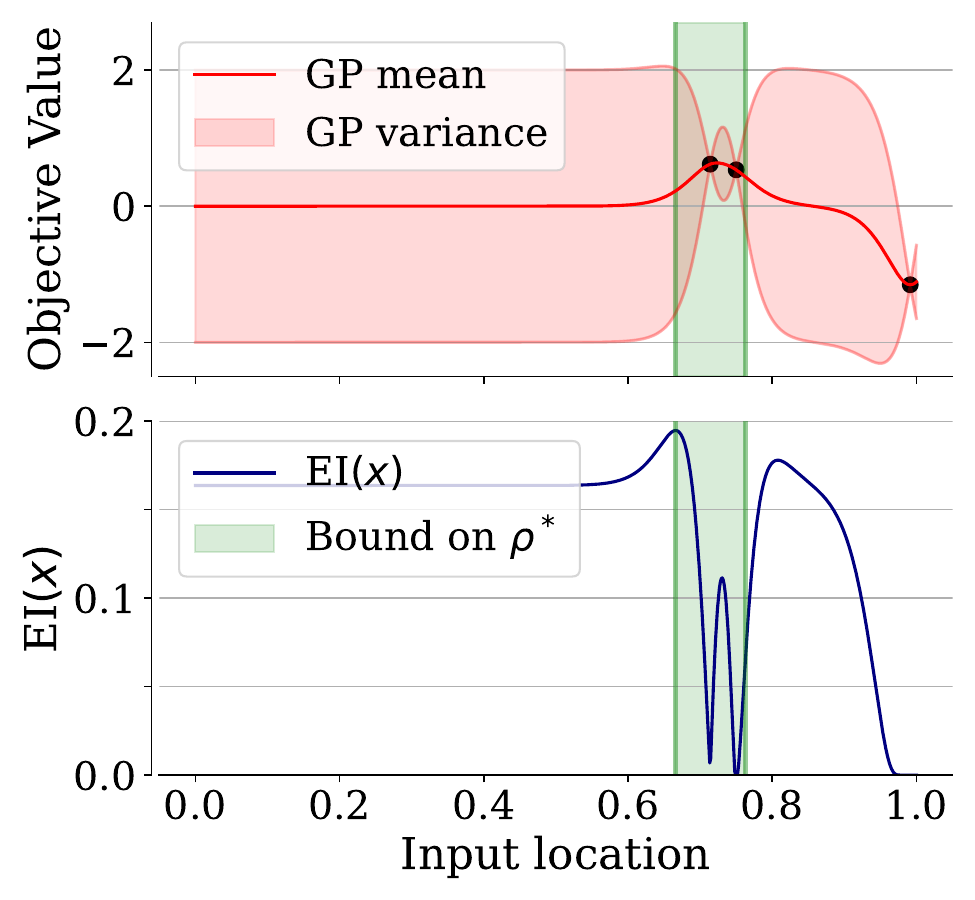}\end{minipage}
\begin{minipage}[b]{0.28\textwidth}
    \includegraphics[width=\linewidth]{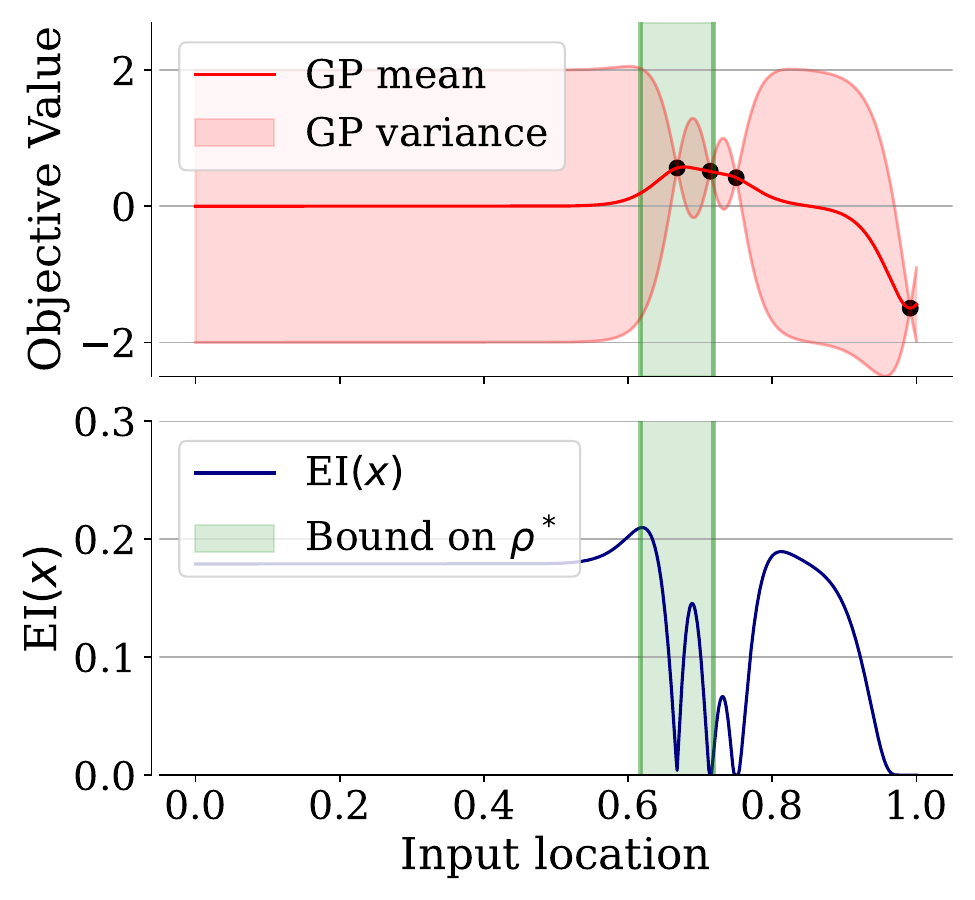}
\end{minipage}\\
\begin{minipage}[b]{0.28\textwidth}
    \includegraphics[width=\textwidth]{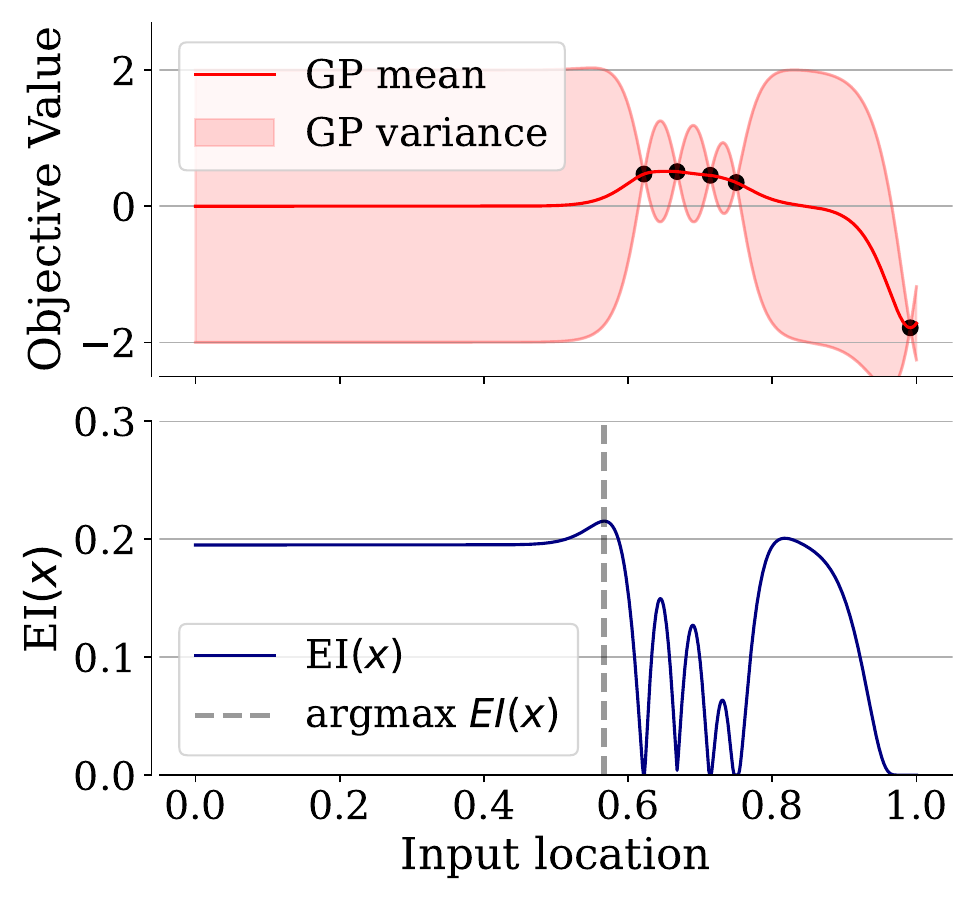}
\end{minipage}
\begin{minipage}[b]{0.28\textwidth}
    \includegraphics[width=\linewidth]{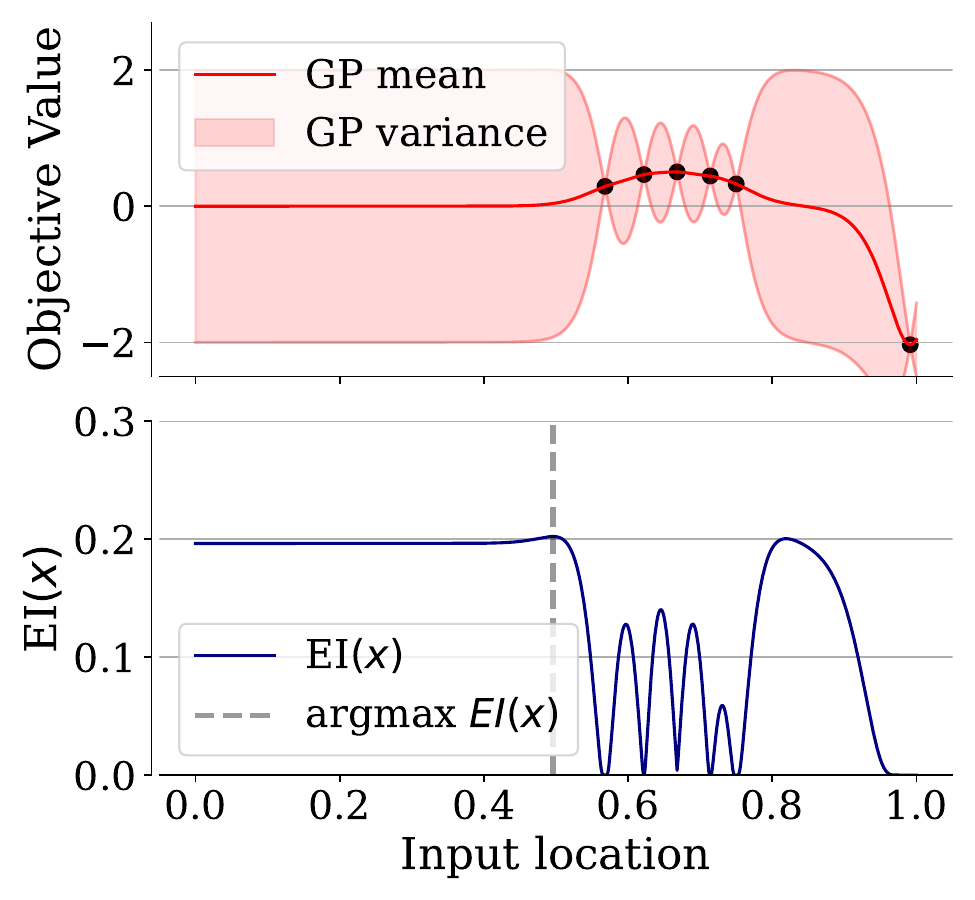}\end{minipage}
\begin{minipage}[b]{0.28\textwidth}
    \includegraphics[width=\linewidth]{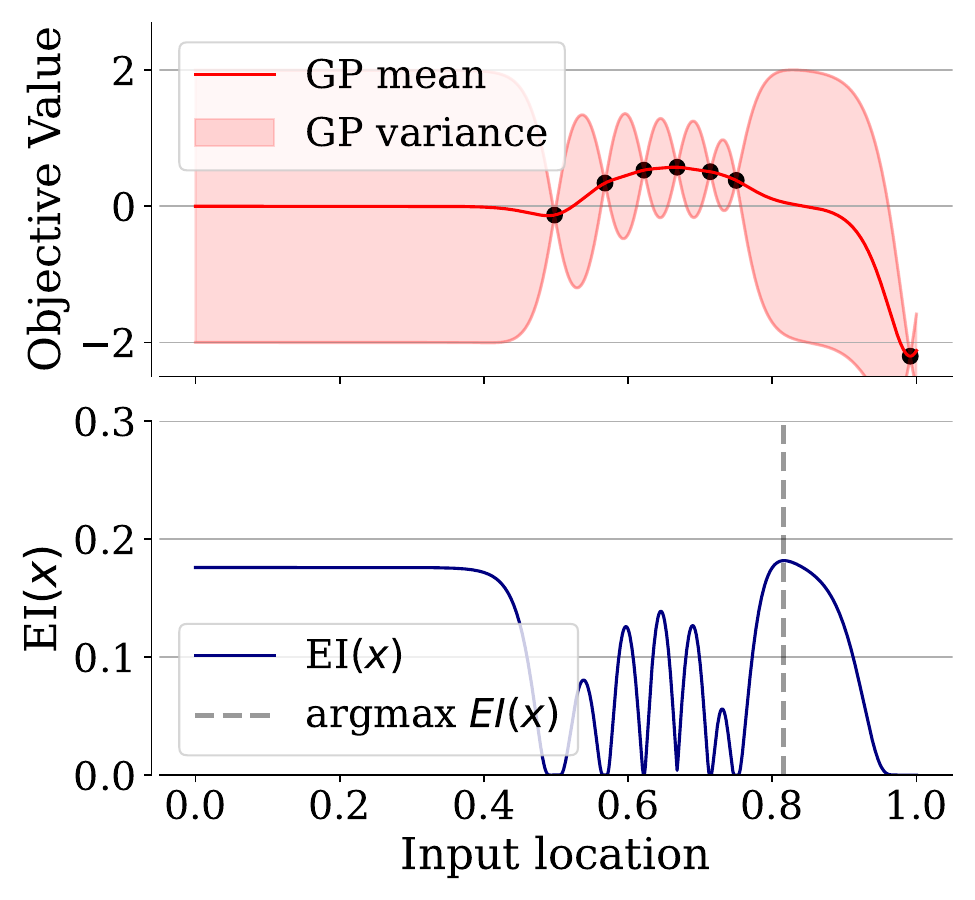}
\end{minipage}
\caption{Visualization of the locality issue for the continuation of Fig.~\ref{fig:bound}, considering the high-complexity model with short lengthscales. For the first three iterations (top row), we visualize the bound on $\rho^*$ since the incumbent is at the leftmost edge of the well-performing region. In the bottom row, the local behavior continues despite no longer observing improvement. Notably, the large variance region in the leftmost part of the search space is never preferred by \ei{} despite continuous re-normalization of the data.}
\label{fig:locality_cont}
\end{figure}

In Fig.~\ref{fig:locality_cont}, we demonstrate the continuation of Fig.~\ref{fig:bound} for six iterations. We visualize the bound on $\rho^*$ in green for the first three iterations (top row) when the incumbent is not in the interior of the good region of the search space. Notably, the large-variance region in the leftmost part of the search space is never considered to be the best, despite repeated re-normalization of the data between iterations.

We re-iterate that the \ei{}-value of the large-variance regions may often be \emph{good, but not optimal}, which is clearly exemplified in Fig.~\ref{fig:locality_cont}.

\subsection{The Boundary Issue through the Lens of Locality}
We now demonstrate how the boundary-seeking behavior~\cite{swersky2017improving} may occur even for a complex model. Importantly, this occurs under special circumstances, when a small number of dimensions account for a large degree of the total model complexity, with other dimensions being practically deactivated. In a high-dimensional setting, where the number of GP lengthscales may be on the same order of magnitude as the number of data points, this becomes increasingly likely.

\begin{figure}[h!]
    \centering
    \includegraphics[width=\linewidth]{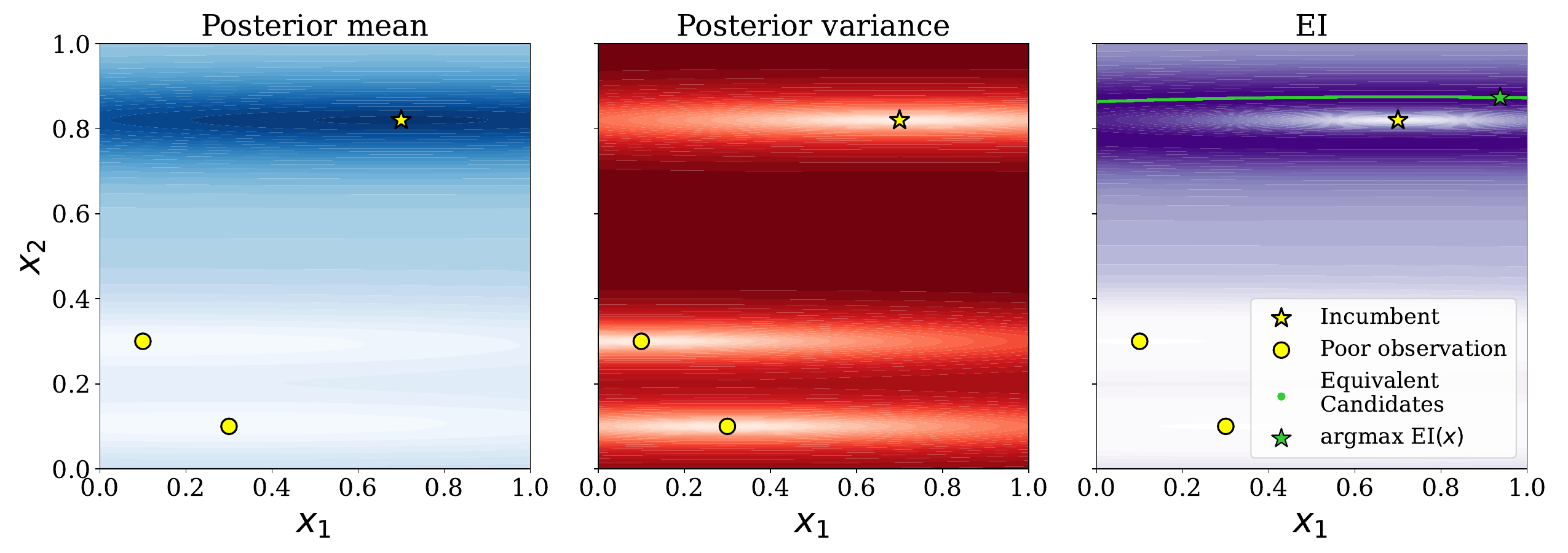}
    \caption{A 2D-visualization of the locality issue for a two-dimensional model of high complexity. The model has a very short lengthscale in $x_2$ ($\ell_1 = 0.1$), and a long lengthscale in $x_1$ ($\ell_2 = 1.75$). From left to right, the plots visualize the posterior mean, variance and \ei{}, where darker colors mean larger values. Close to the incumbent (yellow star), the EI values are high. The green line in the rightmost plot indicate a region of almost equivalent candidates to the $\argmax$ (green star) where EI is within a factor $10^{-5}$ of the $\max$. }
    \label{fig:mle_locality}
\end{figure}

In Fig.~\ref{fig:mle_locality}, we visualize how \ei{} may choose to query high-correlation points that are seemingly distant, and even along the boundary of the search space. We visualize three observations (yellow), the posterior moments, and the value of \ei{} across the search space. The model has a short lengthscale in $x_2$, but the large lengthscale along the $x_1$ dimension allows for traversal across the dimension with very little change in predictions and correlation. For this setting, the optimal query (green star) is close to the incumbent, within our correlation bound. However, all of the green line constitute queries that are within a factor $10^-5$ of the $\max$ of \ei{}. Queries along this line are decidedly \textit{not} exploratory ones (in the classical high-variance sense), since they have very high correlation with the incumbent. However, they cover both boundaries, and a large region outside the bounds of the search space. Due to the instability of MLE in the overparametrized, high-dimensional setting, long-lengthscale may appear frequently dimensions, and can also vary substantially between model fittings.

\begin{figure}[h!]
    \centering
    \begin{minipage}[b]{0.52\textwidth}
    \includegraphics[width=\linewidth]{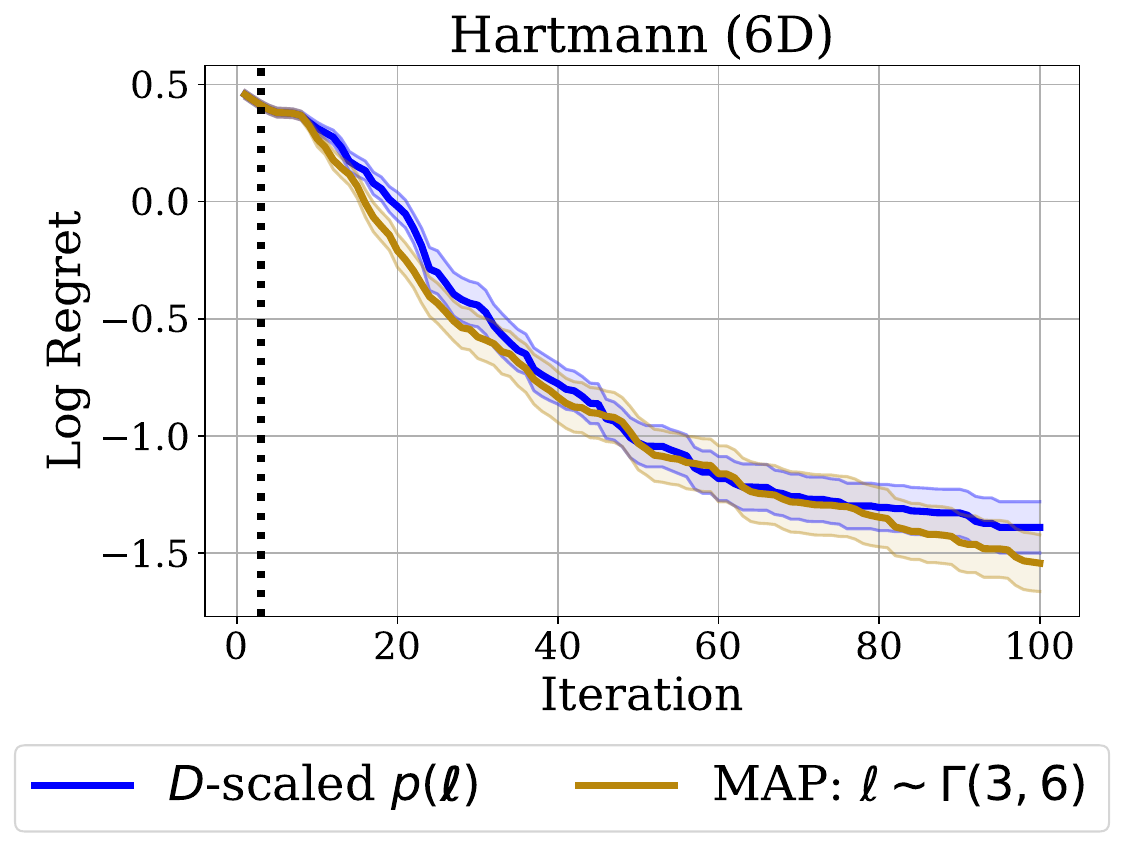}
\end{minipage}
    \begin{minipage}[b]{0.46\textwidth}
    \includegraphics[width=\linewidth]{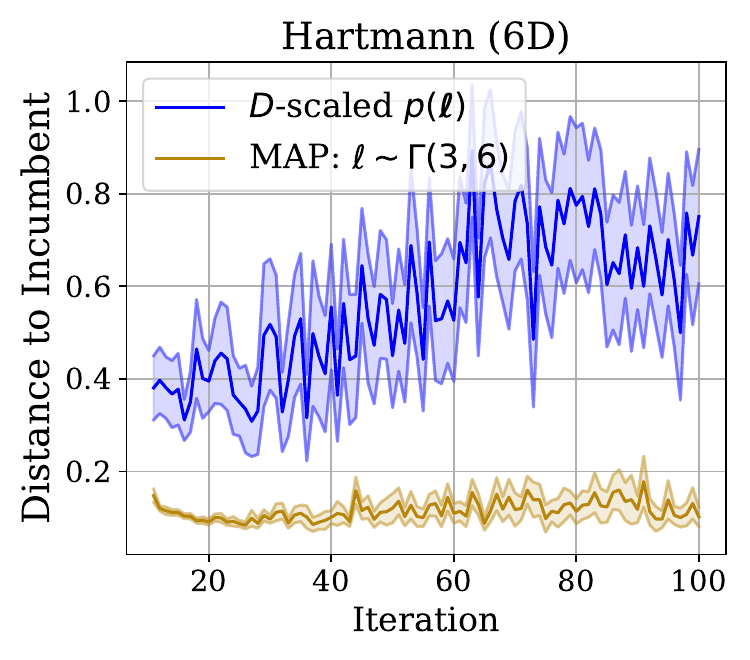}
\end{minipage}
    \caption{(left) Regret performance a short, conventional lengthscale prior (yellow) and the DSP (blue) across 10 repetitions. Conventional MAP and DSP perform comparably in final performance. (right) Average distance to the incumbent for conventional MAP and DSP per iteration, averaged over 20 repetitions. Conventional MAP exemplifies the locality issue, as it almost exclusively queries in close proximity to the incumbent, whereas DSP displays a balance between exploration and exploitation, as visualized by the wide error bars.}
    \label{fig:hartmann}
\end{figure}

Lastly, we demonstrate the drastically different behaviors that may arise even in low-dimensional problems, such as the non-embedded variant of Hartmann (6D). Since our enhancement is applicable to low-dimensional problems as well, we demonstrate its ability to exhibit a well-calibrated exploration-exploitation trade-off. We demonstrate the performance of MLE, conventional~\cite{balandat2020botorch}, low-lengthscale MAP ($\ell_i\sim \Gamma(3, 6)$) and the DSP in Fig.~\ref{fig:hartmann} 
While conventional MAP and DSP achieve almost identical final performance, we now consider the degree to which the locality issue is present for each method by considering the minimal distance to a preceding data point for each query throughout the optimization round. Fig.~\ref{fig:hartmann} visualizes this, and we see that conventional MAP almost never makes an exploratory query after the DoE phase, and takes substantially smaller local steps than our enhancement. While final performance is not negatively impacted, Fig.~\ref{fig:hartmann} reveals that conventional MAP is effectively performing an initial random search followed by 140 iterations of local search. While potent, the high complexity of the model effectively disables the global, exploratory component of the BO algorithm.

\section{Low-budget Performance}~\label{app:anytime}

\begin{figure*}[htbp]
    \centering
    \includegraphics[width=\linewidth]{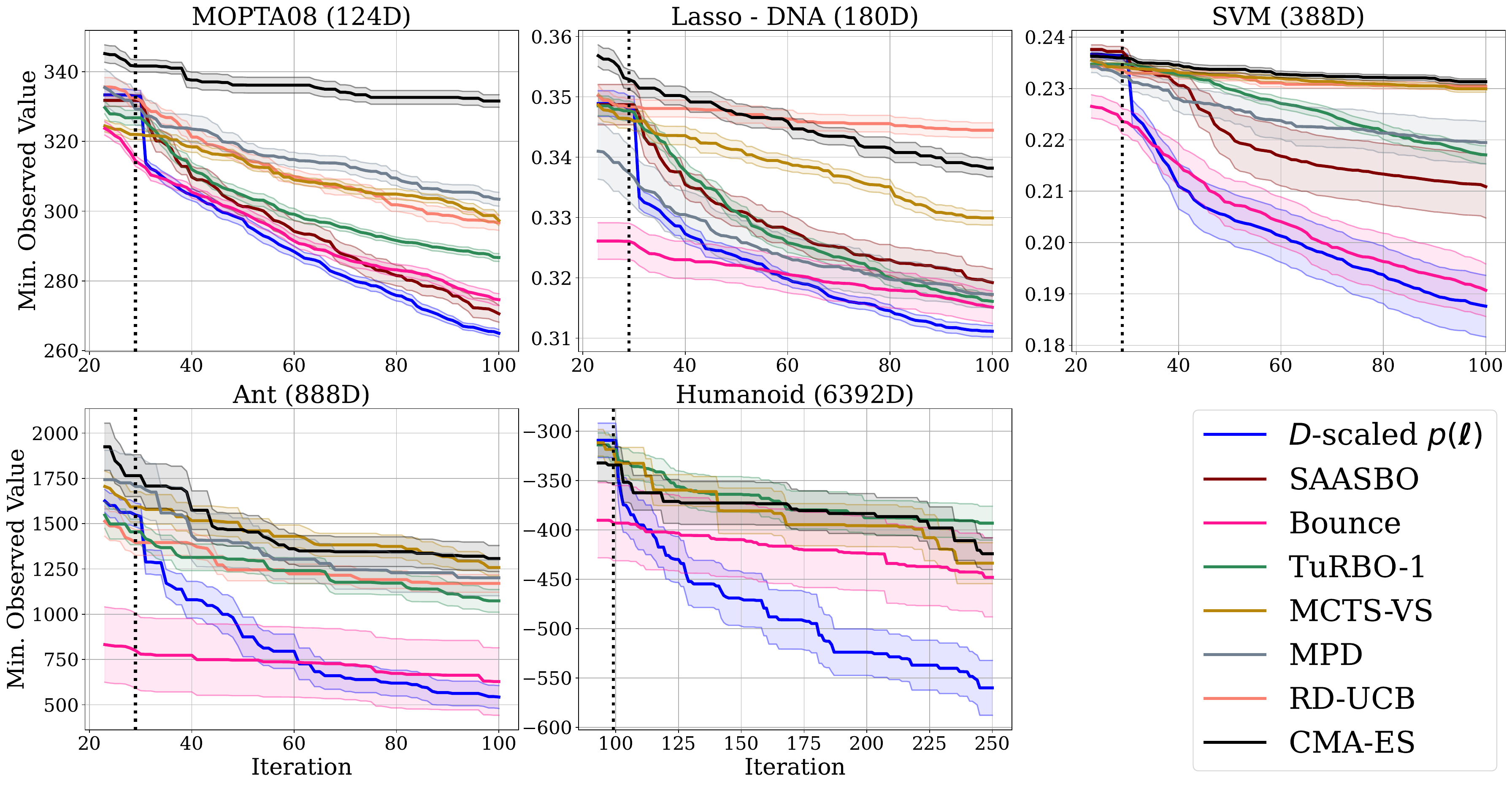}
    \vspace{-1mm}
    \caption{Best observed value of all baselines on five real-world tasks from of various domains across 20 repetitions (10 for SAASBO) and Humanoid. The DSP performs best across all tasks for a low-budget settings, and offers a pronounced initial jump in performance after DoE.}
    \label{fig:lowbudget}
    \vspace{-1mm}
\end{figure*}

We display the performance of the DSP on the real-world benchmarks from Sec.\ref{sec:results}. In Fig.~\ref{fig:lowbudget}, we observe that for all real-world tasks, the DSP provides the strongest initial performance. DSP rapidly improves after initialization, which indicates potent inference ability. Notably, the DSP does not rely on identifying a subset of active variables, yet provides superior performance to the alternatives.
\section{Ablation Studies}\label{app:abl}
We ablate the hyperparameters of the enhanced vanilla BO algorithm. Specifically, we provide ablations for both the assumed complexity and the uncertainty of the complexity measure, as well as comparisons to BO with conventional MAP priors~\cite{balandat2020botorch} of $\ell_i\sim \Gamma(3, 6)$ with high density on low values for the lengthscales, as well as MLE. Lastly, we demonstrate the performance of our enhancement with a learnable signal variance hyperparameter $\sigma_f^2$. We evaluate on a subset of the tasks presented in Sec.~\ref{sec:results}, and include both synthetic, axis-aligned tasks and real-world tasks.

We find that generally, Vanilla BO is highly performant across almost an order of magnitude of complexity, as displayed in Fig.~\ref{fig:abl_comp_syn}, and very robust to changes in uncertainty of the complexity, as shown in Figs.~\ref{fig:abl_uncert_syn} and~\ref{fig:abl_uncert_real}. Assuming too low complexity, however, can occasionally be detrimental to performance, as displayed in Fig~\ref{fig:abl_comp_real}.

\subsection{Conventional MAP and MLE}\label{app:mapmle}
First, we show the performance of our enhancement to vanilla BO compared to MLE and MAP using the default BoTorch~\cite{balandat2020botorch} priors ($\ell_i\sim \Gamma(3, 6)$) on the hyperparameters, which match the description given on conventional priors in Sec.~\ref{sec:bo}. In the higher-dimensional tasks, MLE hyperparameter fitting is very unstable, due to the large number of hyperparameters in relation to the number of observations. Thus, full repetitions of MLE-estimated hyperparameters are not available for many tasks. We run a stabilized variant of MLE fitting as a complement, where the previous iteration's hyperparameters are in case of a crash during model fitting. This variant achieves competitive performance on many tasks, further demonstrating the capabilities of Vanilla Bayesian optimization fro high-dimensional tasks.

We observe that MAP estimation achieves impressive performance for some tasks, even marginally outperforming our method on a few tasks. We note, however, that conventional MAP severely suffers from the locality issue on all tasks, as it simply performs a BO-guided local search around the best point found during DoE. We visualize the locality behavior of all three methods in App.~\ref{app:boundary_emp}.
\begin{figure*}[h!]
    \centering
    \includegraphics[width=\linewidth]{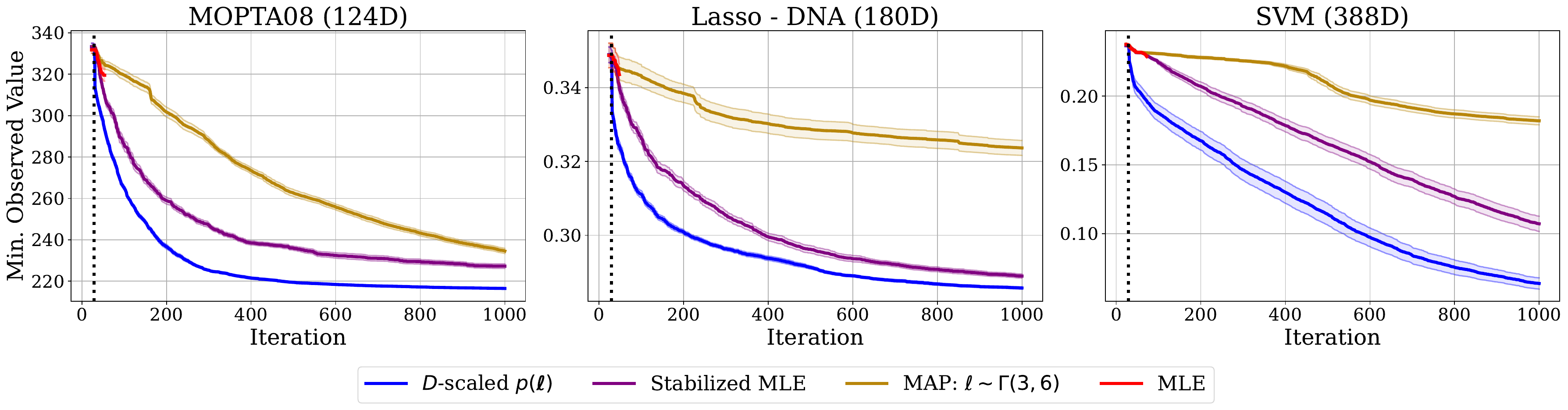}
    \vspace{-4mm}
    \caption{Average log regret of uncertainty ablations on Levy (4D) and Hartmann (6D) 
    synthetic test functions of varying ambient dimensionality for DSP (blue), MLE-estimated hyperparameters (red), stabilized MLE-estimated hyperparameters (purple) and hyperparameters fit through BoTorch default hyperpriors ($\ell_i\sim \Gamma(3, 6)$, green) across 10 repetitions (20 for the default algorithm in blue). $\ell_i\sim \Gamma(3, 6)$ priors perform well on the Hartmann tasks, but searches extremely locally around the best found point from DoE. MLE runs into errors related to fitting instability early on for a majority of all tasks.}
    \vspace{-1mm}
    \label{fig:mlemapsyn}
\end{figure*}

\begin{figure*}[h!]
    \centering
    \includegraphics[width=\linewidth]{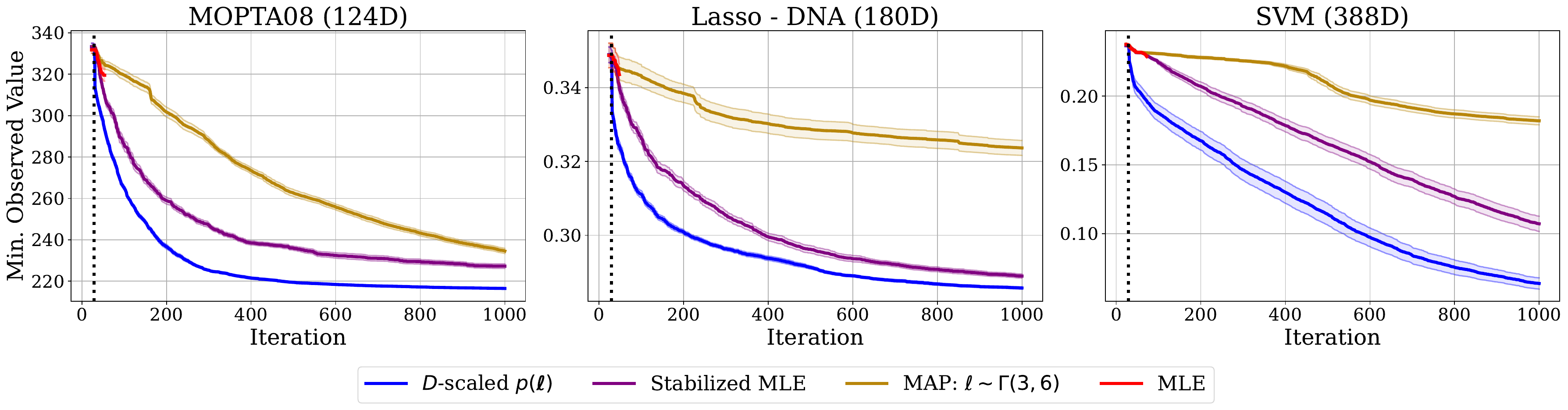}
    \vspace{-4mm}
    \caption{Average minimal observed value on MOPTA, Lasso-DNA and SVM for DSP (blue), MLE-estimated hyperparameters (red), stabilized MLE-estimated hyperparameters (purple) and hyperparameters fit through BoTorch default hyperpriors ($\ell_i\sim \Gamma(3, 6)$, green) across 10 repetitions (20 for the default algorithm in blue). $\ell_i\sim \Gamma(3, 6)$ priors perform well on MOPTA, and MLE runs into errors related to fitting instability early on for all tasks.}
    \vspace{-1mm}
    \label{fig:ablmlemapreal}
\end{figure*}

\subsection{Fixed Complexity - Varied uncertainty}\label{app:ablsigma}
We ablate the uncertainty of the LogNormal prior, keeping the mode fixed. Since the mode of the LogNormal distribution is computed as $\mu - \sigma^2$, we must adjust the $\mu$ term when changing $\sigma$ to keep the mode in place. A larger uncertainty term will increase the ability for Vanilla BO to disregard dimensions that appear unimportant, but also be quicker in identifying dimensions of high importance.
\begin{figure*}[h!]
    \centering
    \includegraphics[width=\linewidth]{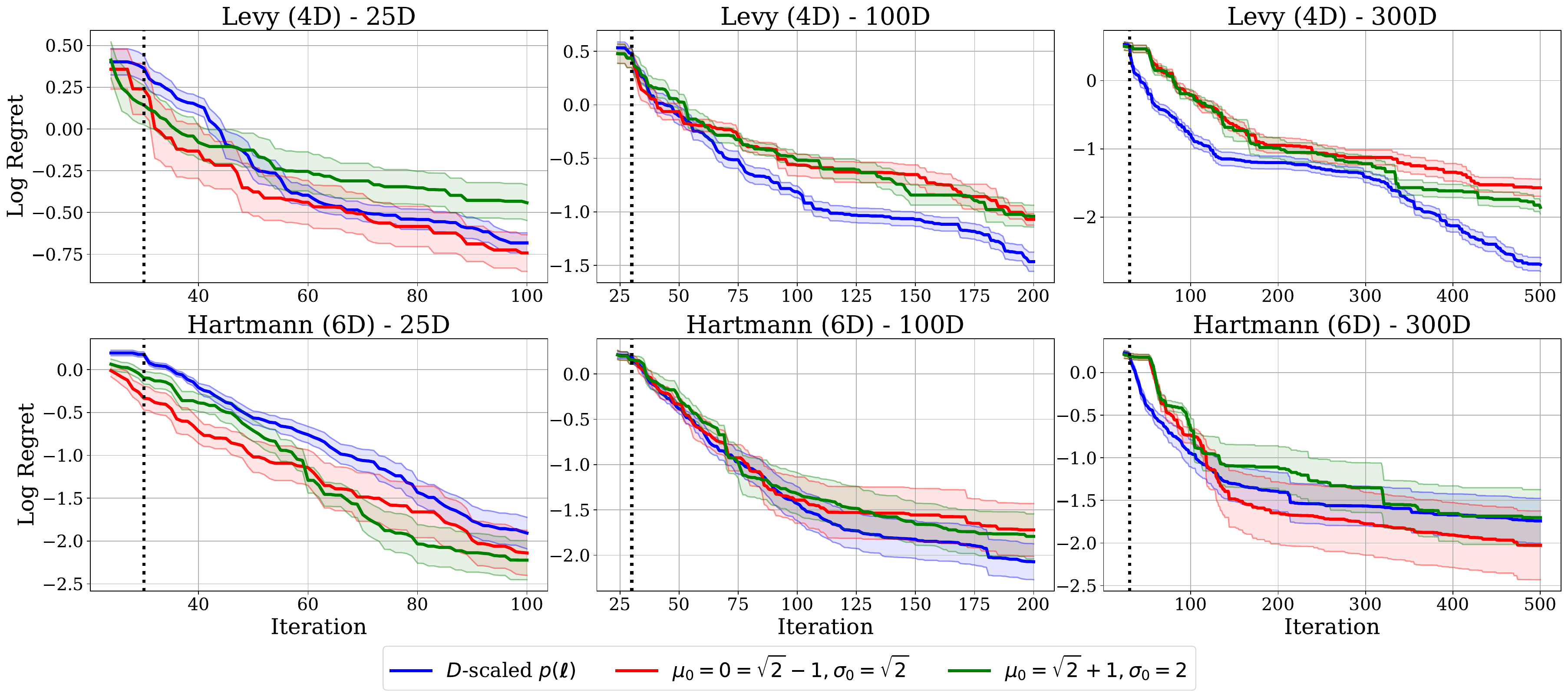}
    \vspace{-4mm}
    \caption{Average log regret of uncertainty ablations on Levy (4D) and Hartmann (6D) 
    synthetic test functions of varying ambient dimensionality across 10 repetitions (20 for the default algorithm in blue).}
    \vspace{-1mm}
    \label{fig:abl_uncert_syn}
\end{figure*}
\begin{figure*}[h!]
    \centering
    \includegraphics[width=\linewidth]{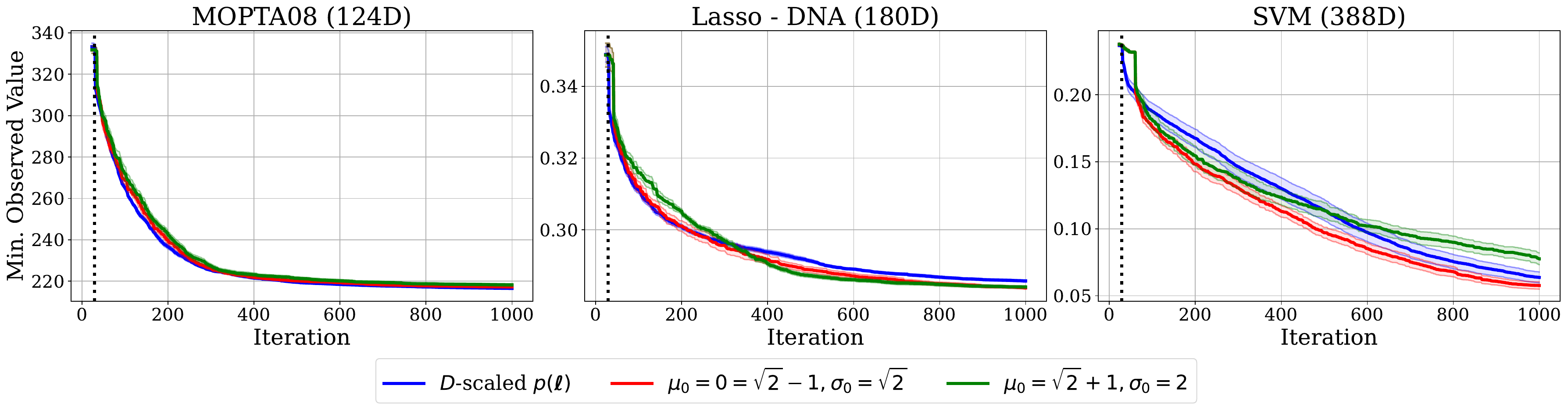}
    \vspace{-4mm}
    \caption{Average minimal observed value on MOPTA, Lasso-DNA and SVM for Vanilla BO with identical modes for the prior, but higher uncertainty (red) and lower uncertainty (green) across 10 repetitions (20 for the default algorithm in blue). Performances are comparable across uncertainty levels, with high uncertainty being marginally worse on SVM.}
    \vspace{-1mm}
    \label{fig:abl_uncert_real}
\end{figure*}
\subsection{Varied Complexity - Fixed uncertainty}\label{app:ablcomp}
We ablate the complexity of the method, keeping the $\sigma$ parameter fixed while varying $\mu$. A larger $\mu$ will increase the default lengthscales, suggesting a lower complexity on the problem at hand. We ablate $\mu$ by $0.5$ in both a positive and negative direction, which corresponds to a factor $1.65$ increase and decrease from the default, respectively.
\begin{figure*}[h!]
    \centering
    \includegraphics[width=\linewidth]{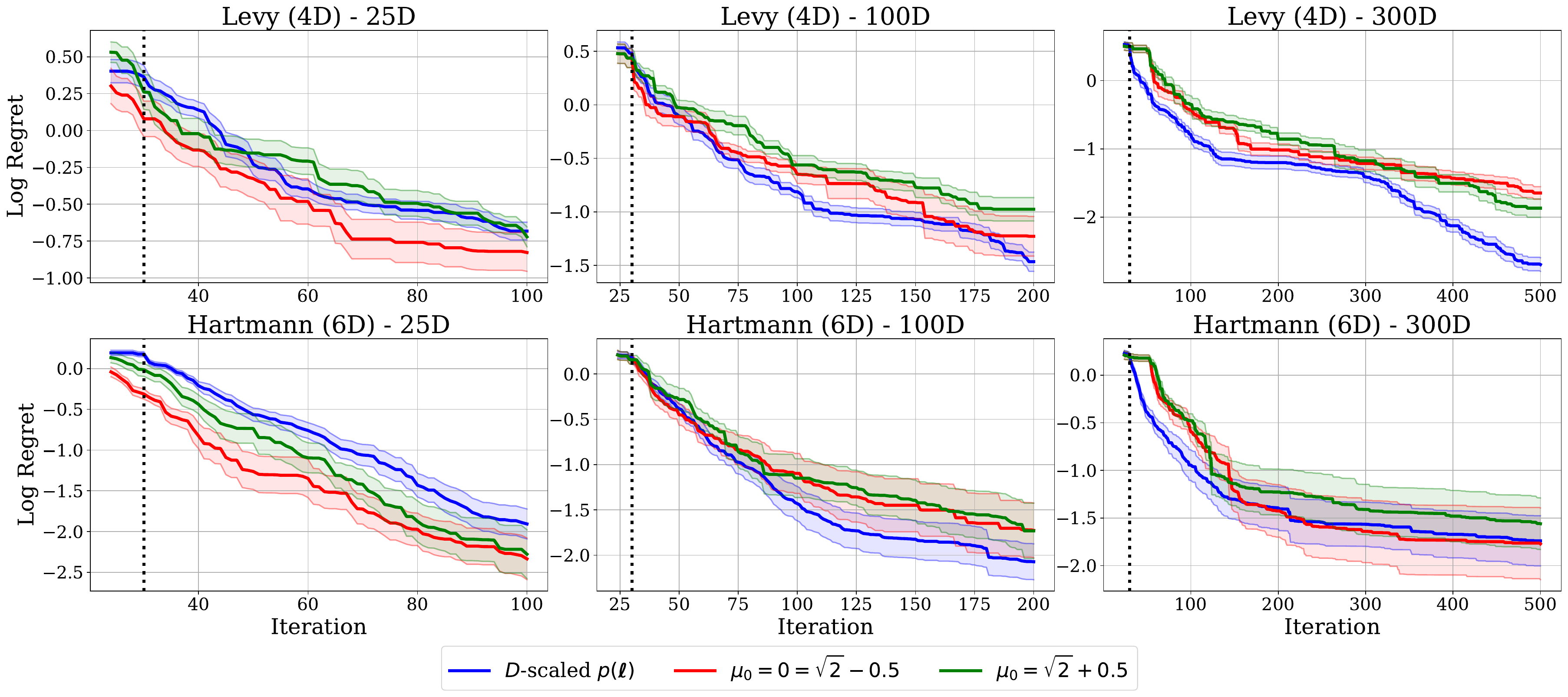}
    \vspace{-4mm}
    \caption{Average log regret of DSP with higher assumed complexity (red) and lower (green) on Levy (4D) and Hartmann (6D) 
    synthetic test functions of varying ambient dimensionality across 10 repetitions (20 for the default algorithm in blue). All instantiations perform comparably, with the low-complexity variant (green) performing slightly worse on average.}
    \vspace{-1mm}
    \label{fig:abl_comp_syn}
\end{figure*}
\begin{figure*}[h!]
    \centering
    \includegraphics[width=\linewidth]{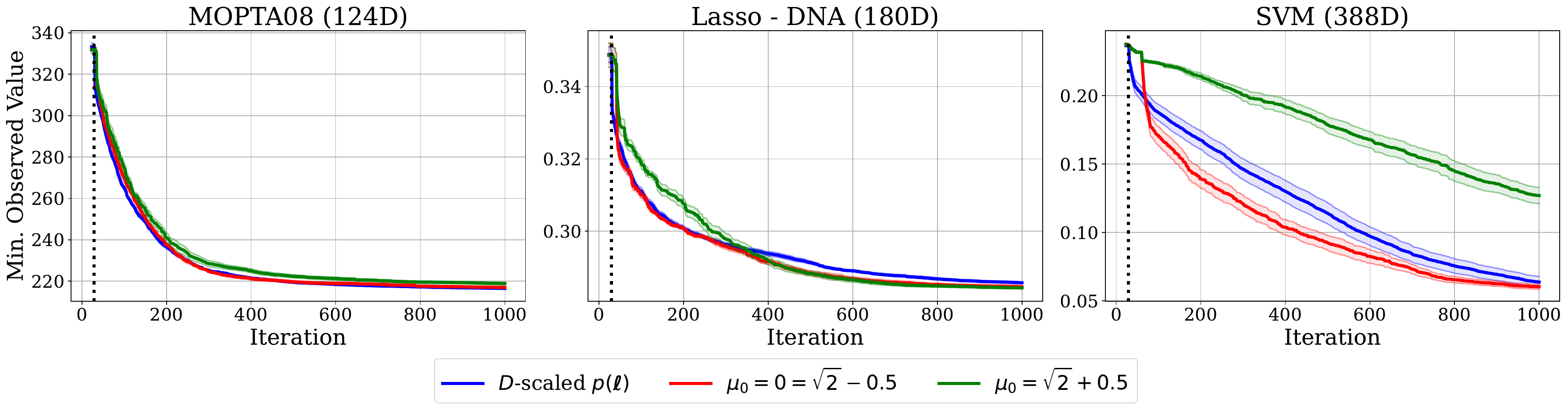}
    \vspace{-4mm}
    \caption{Average minimal observed value on MOPTA, Lasso-DNA and SVM for DSP with higher assumed complexity (red) and lower (green) across 10 repetitions (20 for the default algorithm in blue). Assuming a substantially higher complexity may have detrimental impact on performance, as demonstrated in SVM for the instantiation with the largest $\mu_0$. }
    \vspace{-1mm}
    \label{fig:muabl_real}
\end{figure*}
\section{Learning the signal variance}\label{app:ablops}
\begin{figure}[tb]
    \centering
\begin{minipage}[b]{0.49\linewidth}
    \includegraphics[width=\textwidth]{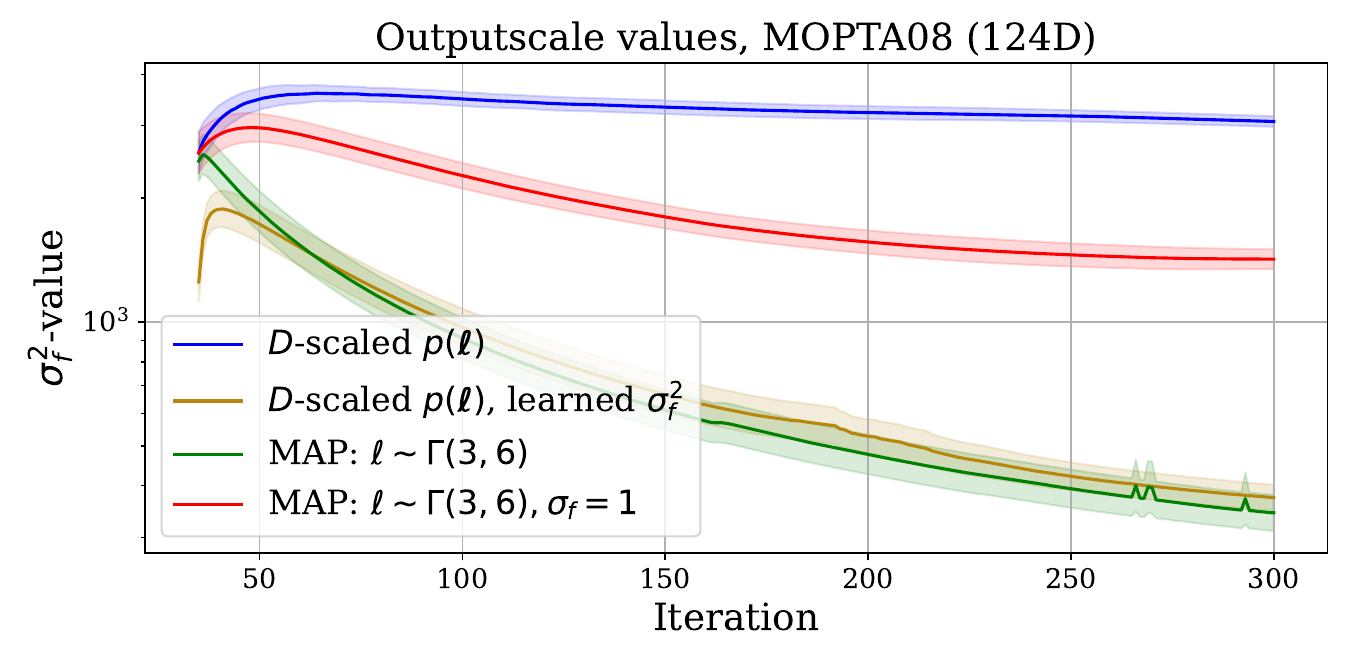}
\end{minipage}
\hfill
\begin{minipage}[b]{0.49\linewidth}
    \includegraphics[width=\linewidth]{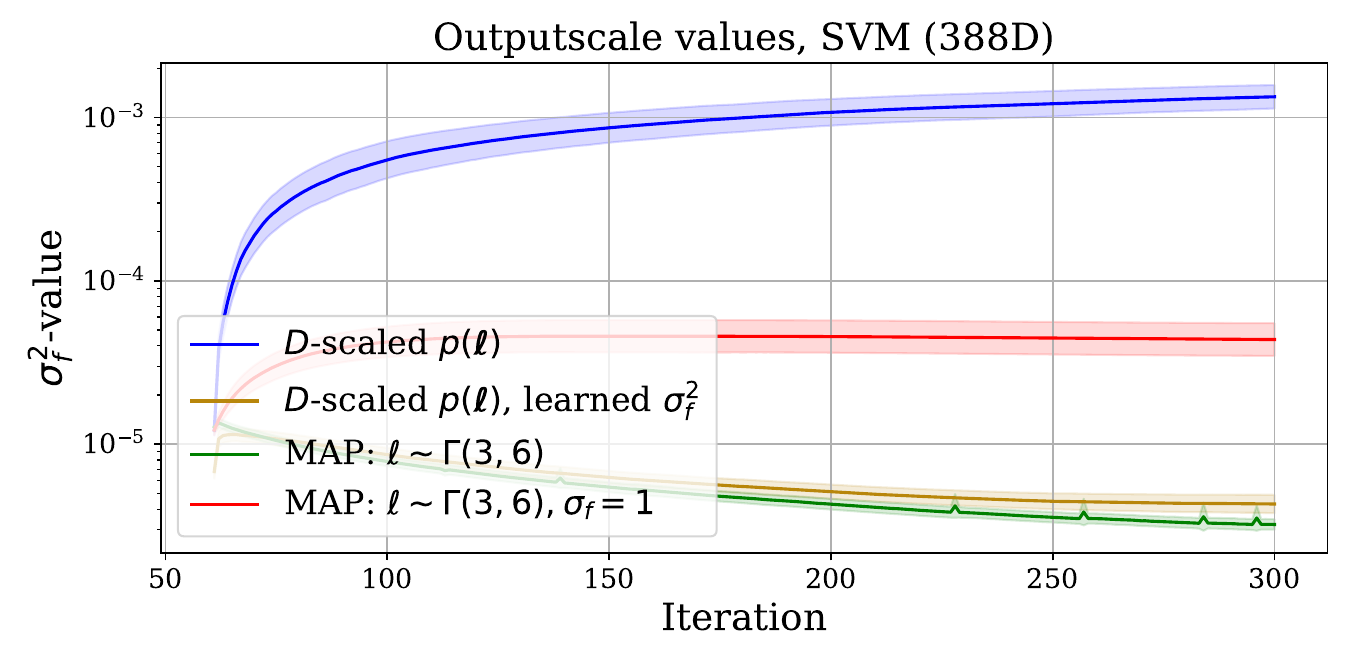}
\end{minipage}
\caption{Outputscale shrinkage on MOPTA08 and SVM for BO (mean and one standard deviation) using four types of models employing either the DSP or conventional MAP, and either fixed or learned outputscale. Learned outputscale results in very pronounced shrinkage over time, which leads to increased exploitative behavior.}
    \label{fig:opsshrink}
\end{figure}

\begin{figure}[tb]
    \centering
\begin{minipage}[b]{0.49\linewidth}
    \includegraphics[width=\textwidth]{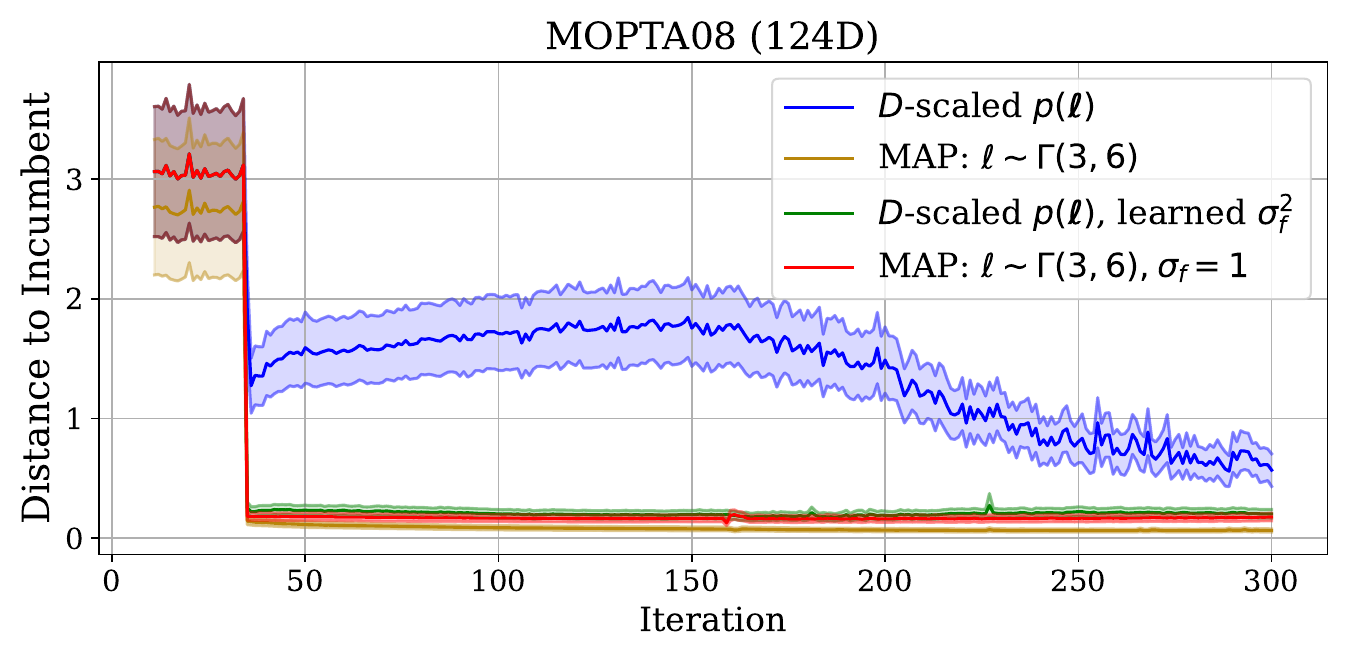}
\end{minipage}
\hfill
\begin{minipage}[b]{0.49\linewidth}
    \includegraphics[width=\linewidth]{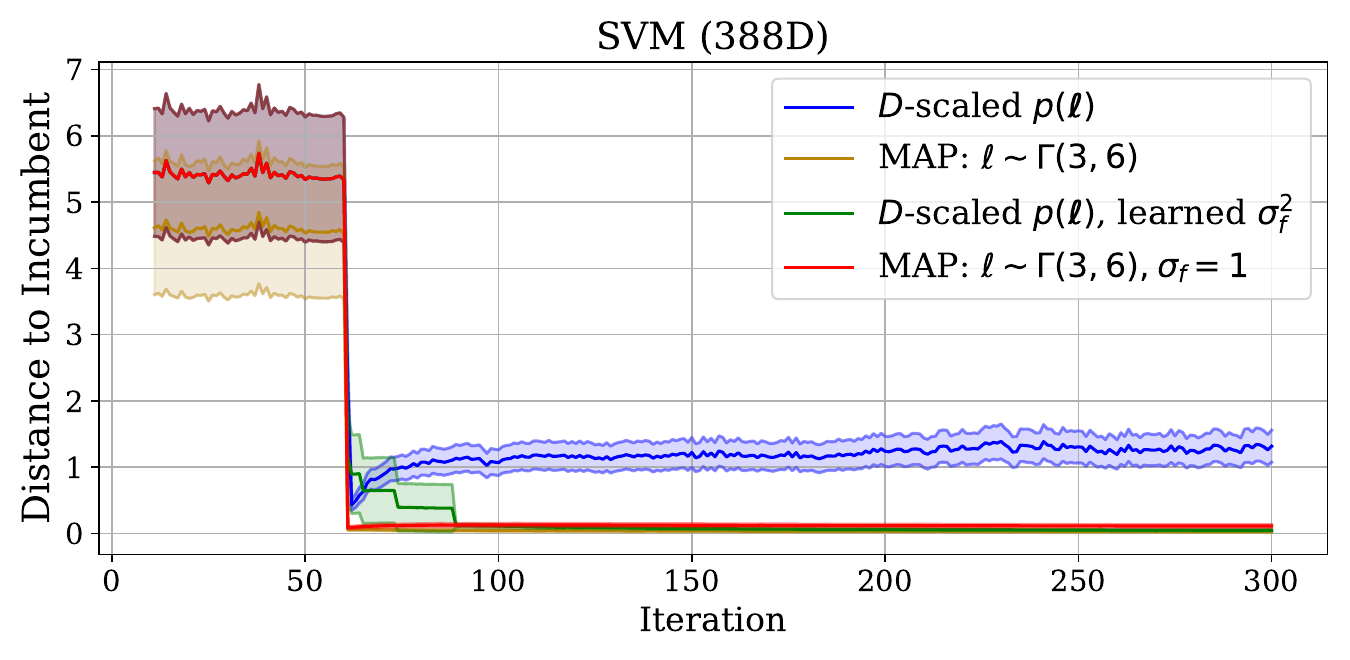}
\end{minipage}
\caption{Average distance from current incumbent (mean and 0.25 standard deviation) on MOPTA08 and SVM for BO using four types of models employing either the DSP or conventional MAP, and either fixed or learned outputscale. Learned outputscale or a conventional prior leads to very local optimization, as the algorithm almost exclusively stays within very close proximity to the current best value.}
    \label{fig:opsincumbent}
\end{figure}

In our method in Sec.~\ref{sec:method}, we propose to not learn the signal variance to safeguard against possible degeneracies related to the active acquisition of data. We ablate the choice here, and demonstrate that, while the performance of a model with a learned signal variance is marginally less stable, the performance is comparable to one where the signal variance is not learned. In Fig.~\ref{fig:opsshrink}, we display how the outputscale hyperparameter shrinks throughout optimization for models with a learned outputscale. Notably, the fixed outputscale variants (red, blue) do not display constant outputscales due to varying data normalization over time. Moreover, in Fig.~\ref{fig:opsincumbent}, we show the exaggerated local search-like behavior that follows from either a learned outputscale or conventional lengthscale prior, as measured by the average distance between the current best point and the upcoming query. The initial samples of high variance and large distance between queries indicates the initial design of experiments via SOBOL sampling.

\subsection{Learning $\sigma_f^2$}~\label{app:opslearn}

We empirically demonstrate the behavior of the model when learning the signal variance for two tasks, MOPTA08 and SVM, for four types of models throughout the BO loop: DSP with fixed $\sigma_f^2$, DSP with learned $\sigma_f^2$, and a conventional $\Gamma(3, 6)$ prior with and without learned $\sigma_f^2$.

\subsection{Performance Ablation on Learned $\sigma_f^2$}~\label{sec:opsperf}
We display the performance of the DSP with and without learned signal variance for a subset of the synthetic and real-world tasks presented in the paper. While the performance is occasionally competitve, the performance is substantially less stable than for the fixed outputscale variant.
\begin{figure*}[h!]
    \centering
    \includegraphics[width=\linewidth]{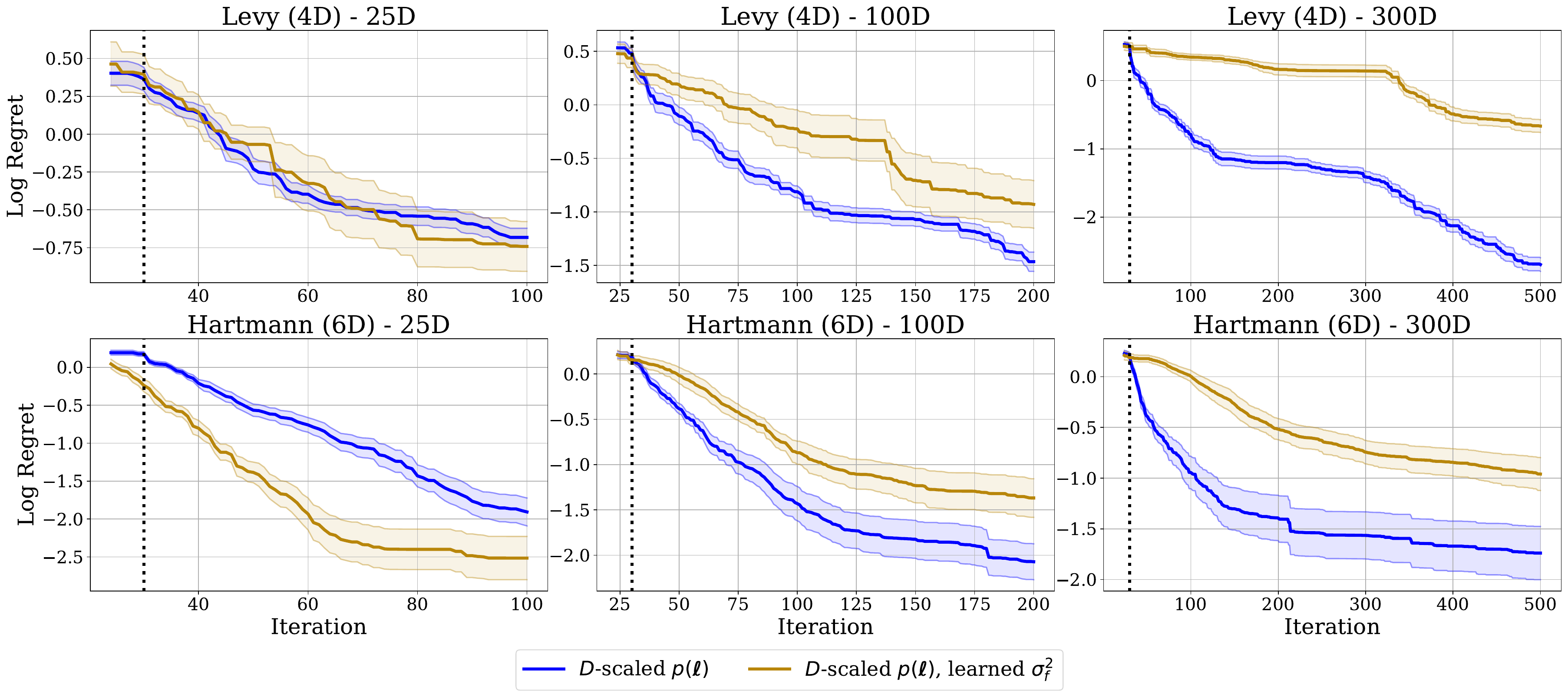}
    \vspace{-4mm}
    \caption{Average log regret of Vanilla BO with a learned signal variance (yellow) and the default algorithm (blue) on Levy (4D) and Hartmann (6D) 
    synthetic test functions of varying ambient dimensionality across 10 repetitions (20 for the default algorithm in blue). The learned signal variance variant performs comparably on some tasks, but displays worse consistency across tasks.}
    \vspace{-1mm}
    \label{fig:abl_opsreal}
\end{figure*}
\begin{figure*}[h!]
    \centering
    \includegraphics[width=\linewidth]{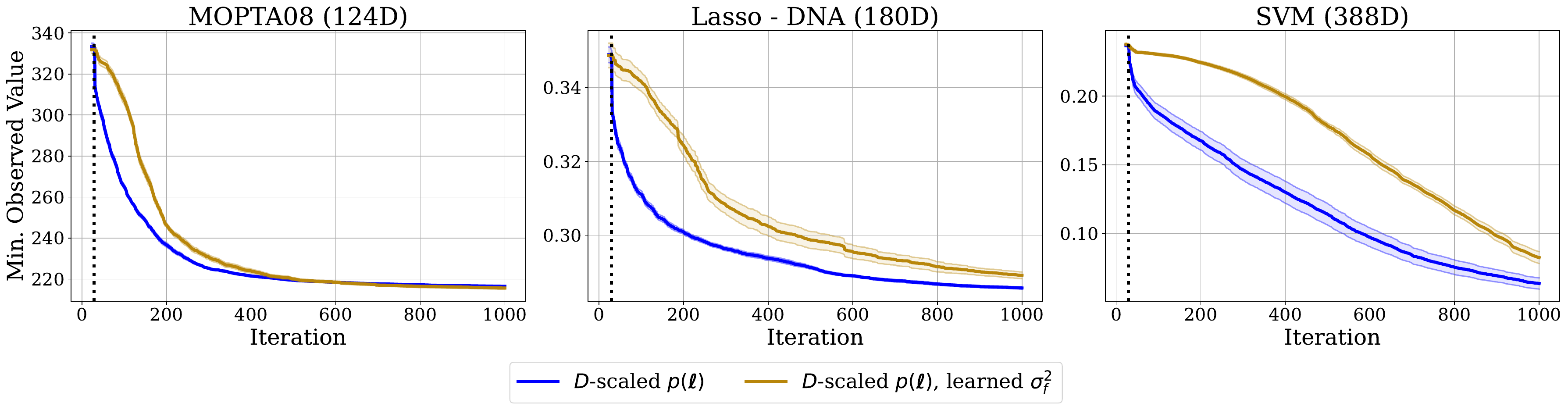}
    \vspace{-4mm}
    \caption{Average minimal observed value of Vanilla BO with a learned signal variance (yellow) and the default algorithm (blue) on MOPTA, Lasso-DNA and SVM across 10 repetitions (20 for the default algorithm in blue). The learned signal variance variant performs comparably on some tasks, but displays worse consistency across tasks.}
    \vspace{-1mm}
    \label{fig:abl_comp_real}
\end{figure*}
\end{document}